\documentclass[preprint,review,10pt]{elsarticle}
\usepackage{tikz}       
\usetikzlibrary{positioning}
\usetikzlibrary{calc}
\usetikzlibrary{shapes.multipart}
\usetikzlibrary{shapes.geometric}
\usetikzlibrary{decorations.pathreplacing,calligraphy}
\usetikzlibrary {decorations.markings} 
\usetikzlibrary{calc}

\usepackage{graphicx}
\usepackage{subcaption}
\usepackage{amsmath}
\usepackage{amsfonts}
\usepackage{multirow}
\usepackage{tabularx}
\usepackage{pifont}
\newcommand*{\barre}[1]{$\overline{\text{#1}}$}

\usepackage{amssymb}
\usepackage{amsmath}
\usepackage{hyperref}

\usepackage{amsthm}

\usepackage{lineno}

\journal{Pattern Recognition}

\begin{document}

\begin{frontmatter}



\title{Local and Global Graph Modeling with Edge-weighted Graph Attention Network for Handwritten Mathematical Expression Recognition} 
\author{Yejing XIE\fnref{label2}}
\author{Richard Zanibbi\fnref{label3}}
\author{Harold Mouchère\fnref{label2}}
\affiliation[label2]{organization={Nantes Universite, Ecole Centrale Nantes, CNRS, LS2N, UMR 6004},
            city={Nantes},
            postcode={F-44300}, 
            country={France}}
\affiliation[label3]{
    organization={Document and Pattern Recognition Lab, Rochester Institute of Technology},
    city={Rochester},
    state={New York},
    country={USA}
}

\begin{abstract}
In this paper, we present a novel approach to Handwritten Mathematical Expression Recognition (HMER) by leveraging graph-based modeling techniques. We introduce a End-to-end model with an Edge-weighted Graph Attention Mechanism (EGAT), designed to perform simultaneous node and edge classification. This model effectively integrates node and edge features, facilitating the prediction of symbol classes and their relationships within mathematical expressions. Additionally, we propose a stroke-level Graph Modeling method for both local (LGM) and global (GGM) information, which applies an end-to-end model to Online HMER tasks, transforming the recognition problem into node and edge classification tasks in graph structure. By capturing both local and global graph features, our method ensures comprehensive understanding of the expression structure. Through the combination of these components, our system demonstrates superior performance in symbol detection, relation classification, and expression-level recognition.
\end{abstract}




\end{frontmatter}


\section{Introduction}
\label{intro}


Mathematical Expression (ME) is a fundamental component of scientific research, engineering development, basic education, and various other fields.
In contrast to more structured, but less intuitive, ME editing tools and markup languages (such as \LaTeX), handwritten mathematical expressions offer greater ease of use for humans but pose a greater challenge for machine recognition due to variations in individual writing styles and writing habits.
Handwritten Mathematical Expression Recognition (HMER), which involves converting handwritten math into markup language for easier computer processing and rendering, is a challenging promising field with various of potential applications.
Compared to Optical Character Recognition (OCR), recognizing handwritten manuscripts is more challenging due to the wide variation in handwriting styles. HMER not only faces the common challenges of handwriting recognition but also has to deal with the added complexity of interpreting the 2D structure of mathematical expressions.
According to different processing objective, HMER can be categorized into Online HMER and Offline HMER. 
Online HMER processes a sequence of temporal trajectories captured by digital devices like tablets and digital pens. Online data is segmented into individual strokes based on pen-down and pen-up interruption. 
While offline expressions are static images collected by scanner, camera or smartphone. 

Existing deep learning architectures for HMER are typically based on encoder-decoder models that process either online or offline inputs to generate \LaTeX{} markup strings. These sequence-to-sequence models, however, fail to leverage the graph structure inherent in mathematical layouts, making it difficult to capture and use the relationships between symbols.
Some algorithms attempt to incorporate graph or tree structures as assistive information for recognition. However, these approaches operate in a latent space that is not directly tied to the input data, such as individual strokes.

In response to this limitation, we aim to explore a graph-based representation of handwritten mathematical expressions (HME) and leverage this structured data for end-to-end stroke-level recognition for online data. This approach seeks to directly model the relationships between strokes, enhancing the recognition process.
In this study, we focus on Online HMER, and the key contributions of this paper are as follows:
\begin{itemize}
    \item[$\bullet$] We propose a novel \textbf{End-to-end model} with an \textbf{Edge-weighted Graph Attention Mechanism (EGAT)}, designed for simultaneous node and edge classification tasks. This model effectively aggregates embedded node and edge features of graph structures, enabling the prediction of node and edge attributes in a single pass.
    \item[$\bullet$] We introduce stroke-level \textbf{Local Graph Modeling (LGM)} and \textbf{Global Graph Modeling (GGM)} that applies the general EGAT model for Online HMER tasks, transforming the Online HMER problem into a graph classification task. Both local and global graph features are captured in this process.
    \item[$\bullet$] We propose a detailed evaluation of our system which achieves strong performance in symbol detection, relation classification, and overall expression level recognition.
\end{itemize}

\begin{figure}
    \centering
    \includegraphics[trim={60 100 60 100},clip,width=0.9\textwidth]{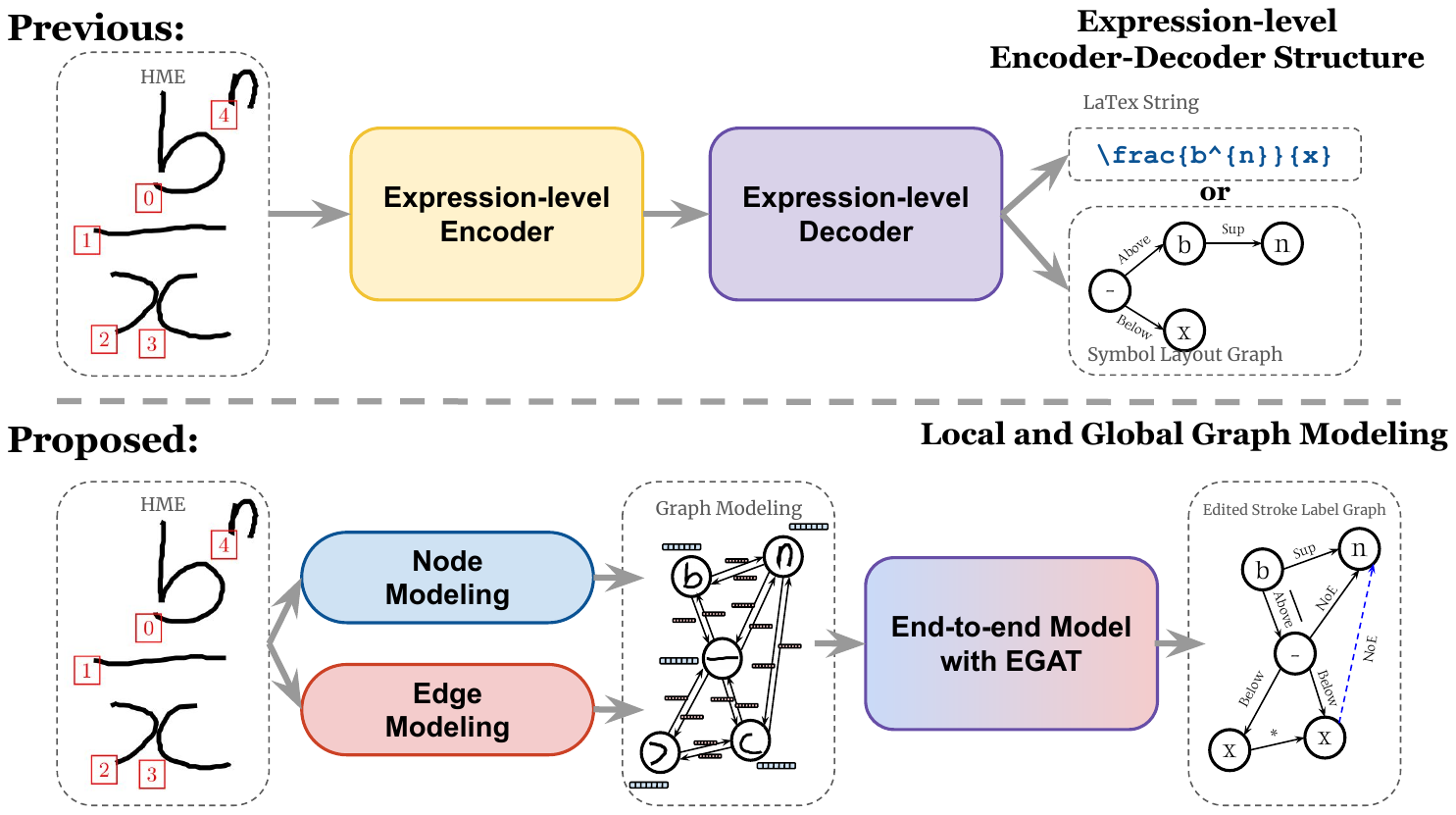}
    \caption{The overview of the proposed end-to-end model. In constrast to encoder-decoder structures, our approach directly aligns the strokes of Online HME with their corresponding final graph representation. of the expression.}
    \label{fig:overview}
\end{figure}

As shown in Fig.\ref{fig:overview}, the proposed system’s general structure is designed to directly align the strokes of Online HME with their corresponding symbol labels, and to map the relationships between strokes to the corresponding relation labels. 

\section{Related Work}
\label{sota}

In the early years, the HMER algorithms are mainly the Structural Recognition Methods, which construct a graph or tree structure with explicit relations between symbols, and then match the graph or tree structure with the predefined rules.
The limitation of structural recognition methods is that the entire recognition is always separated into several sub-tasks, such as symbol segmentation, symbol classification, relation classification and structural analysis, which are executed independently and later steps are based on the results of the previous steps.

Grammar-based methods and graph-based methods are both structural recognition methods.
Same with the other string-based languages, the expression grammar which is a set of rules for the expression structure, is also introduced into the HMER, such as the context-free grammar (CFG) utilized by early HMER systems \cite{yamamoto2006line,lavirotte1998mathematical,AWAL201468,chou1989recognition}. 
Graph-based methods take the advantage of the graph structure of the expression, which can effectively represent the relations between symbols, such as the Stroke Label Graph (SLG).
According to different probability calculation system by the features of expression, graph-based parsing strategies such as minimum spanning tree (MST) can be applied to the determination of the final expression structure, such as \cite{matsakis1999recognition,koschinski1995segmentation,winkler1997online,hu2016mst}

Nowadays, HMER is more and more based on the deep learning methods with the help of high performance computing resources.
Deep learning methods reduce the dependency of hand-crafted features and also grammar rules, significantly improve the recognition accuracy. 
In addition, the structural-based, grammar-based and also graph-based theories are also introduced into the deep learning models, which can improve the interpretability of the models.

In general, Encoder-Decoder structure is the most popular deep learning model for HMER, Sequence to Sequence (Seq2Seq) model including off-line image to \LaTeX{} string and online traces to \LaTeX{} string, is the major trend in the HMER field.
Track, Attend and Parse (TAP) \cite{zhang2018track} and Watch, Attend and Parse (WAP) \cite{zhang2017watch} are basic Seq2Seq models for online and offline HMER, which applied Gated Recurrent Unit (GRU) and Fully Convolutional Network (FCN) as the encoders, while both of their decoders are based on another GRU. 
TAP and WAP popularized the Encoder-Decoder structure in the HMER field, and also the attention mechanism, which can effectively capture the relations between symbols.
With the popularity of Transformer \cite{vaswani2017attention}, Transformer network widely applied as the Decoder \cite{zhao2021handwritten}. \cite{tang2024offline} constructed graph structure with the help of YOLOv5 symbol detector and visual feature extractor, and then applied Graph Attention Network as Encoder and Transformer as Decoder for the recognition.
Some deep learning strategies are employed to improve the models' performance, such as \cite{wu2020handwritten}, who proposed paired adversarial learning to improve the discrimination of Encoder-Decoder structure. A novel Attention Refinement Module (ARM) is proposed by \cite{zhao2022comer} to adopts the coverage information in Transformer-based Decoder. 
\cite{lin2022cclsl} proposed an Encoder-Decoder structure with contrastive learning to learn semantic-invariant features between printed and handwritten symbols.
Syn\-tax-aware network (SAN) \cite{yuan2022syntax} combined syntax information of math expression with Encoder-Decoder structure, which is a novel incorporation of grammar-based methods and deep learning methods. 

Alternatively, some models take the advantage of the tree structure of math expressions, applied a Tree-based Decoder for better representation of the expression structure.
\cite{zhang2017tree} extended chain-structured BLSTM as tree structure for Online HMER, while \cite{truong2021learning} proposed a symbol-relation temporal classifier with BLSTM and also a series of handmade path extraction rules for tree reconstruction.
A sequential relation decoder was proposed to decode the encoded online trajectory \cite{zhang2020tree} and offline images \cite{peng2021image} into tree structure. 
However, these tree based decoders always convert tree structure by a fixed order, thus they fail to take full advantage of the diverse expressions of tree. 
To improve on this point, C. Wu et al. \cite{wu2022tdv2} proposed a novel tree decoder (TDv2) to make maximum advantage of tree structure labels.

With the development of graph theory in deep learning, the graph-based methods are also introduced into the HMER field, which blurred the importance of syntactic correctness in the output expression structure, aimed to obtain high-dimensional structural features with the help of deep neural networks.
\cite{mahdavi2020visual} adopted visual parsing, especially graph parsing for HMER and proposed Query-Driven Global Graph Attention (QD-GGA) algorithm. 
\cite{wu2021graph} proposed a Graph-to-Graph model for Online HMER with encoder-decoder structure, which convert the online trajectory into symbol level graph and lack the direct correspondence between the input and output.
\cite{tang2022offline} proposed an offline HMER algorithm with Graph Reasoning Network (GRN). 

Building on recent works in HMER, where graphs are increasingly used to represent the structural relationships between handwritten strokes and symbols, Graph Neural Networks (GNNs) \cite{scarselli2008graph} have emerged as a powerful tool for handling such graph-structured data.
GNNs have gained significant attention in recent years due to their ability to model and process data represented as graphs. A key mechanism in GNNs is message passing \cite{gilmer2017neural}, where nodes in the graph exchange information with their neighbors. Each node aggregates messages from its connected nodes to update its own features, allowing the model to capture both local and structural information. An extension of this is the Graph Attention Network (GAT) \cite{velivckovic2017graph}, which improves the message-passing process by assigning different attention weights to each edge, allowing the model to focus more on important relationships between nodes. Additionally, the concept of a master node \cite{xu2018powerful} can be introduced in graph models to enhance the capture of global information. The master node is connected to all other nodes, aggregating the information from the entire graph and enabling a more comprehensive understanding of the graph structure. This combination of message passing, attention mechanisms, and master nodes allows GNNs to excel at learning complex patterns in graph-based data, which is the basement of our proposed method.
\section{Methodology}
\label{method}

The approach outlined in this paper is composed of two key components, as shown in Fig. \ref{fig:overview}.
The first component is a lightweight End-to-end model with an Edge-weighted Graph Attention Mechanism (EGAT), which is designed for simultaneous node and edge classification tasks.
This model effectively aggregates embedded node and edge features of graph structures, enabling the prediction of node and edge attributes in a single pass, which will be detailed in section \ref{method:end2end}.
The second component is a stroke-level Graph Modeling method (LGM and GGM), which applies the general EGAT for Online HMER task, transforming the Online HMER problem into a node and edge classification tasks in graph structure.
Both local and global graph features are captured in this process, which will be detailed in section \ref{method:modelisation}.

\subsection{End-to-end Model with Edge-weighted Graph Attention Mechanism}
\label{method:end2end}
With the excellent graph structure modeling capability and also the efficient relations and dependencies caption ability, GNNs can be employed successfully in well-modeled graph. 
In addition to the stroke information embedded in nodes, the relations between nodes embedded in edges are also important for some structural recognition tasks, such as HMER. 
However, edge information can not be represented as reachability or a simple weight, and is supposed to be dynamically involved in message passing and features update.
Thus, we propose an end-to-end model with node and edge embedding modules, edge-weighted graph attention modules for message passing and feature integration, and readout modules for node and edge classification tasks.
The model performs a graph-to-graph transformation, which refers to the conversion of the modeled graph $\mathbf{G}$, containing both node and edge features, into the ground truth graph $\mathbf{\hat{G}}$ with node and edge labels.

\begin{equation}
    \mathbf{G} = \left(\mathbf{V},\mathcal{A}, \mathbf{H}, \mathbf{B}\right)
    \label{eq:LGM-EGAT}
\end{equation}

As shown in Eq.\ref{eq:LGM-EGAT}, a general graph is defined by 4 parts. 
$\mathbf{V} = \{\mathbf{v}_1,\mathbf{v}_2, \ldots, \mathbf{v}_n\}$ is the set of nodes in the graph, with $n$ representing the number of nodes.
$\mathcal{A} \in \{0,1\}^{n \times n}$ is the adjacency matrix, where each element $\mathcal{A}_{ij}$ represents the presence or absence of an edge between nodes $\mathbf{v}_i$ and $\mathbf{v}_j$.
\begin{equation}
    \mathcal{A}_{ij} = \begin{cases}
        1, & \text{if there is an edge between $\mathbf{v}_i$ and $\mathbf{v}_j,$} \\
        0, & \text{otherwise.}
    \end{cases}
\end{equation}
The embedded node features of all the graph $\mathbf{H} \in \mathbb{R}^{n \times d_{1}}$, where $\mathbf{h}_i \in \mathbb{R}^{d_{1}}$ is the embedded feature of node $\mathbf{v}_i$, and $d_{1}$ is the dimension of node features.
The embedded edge features $\mathbf{B} \in \mathbb{R}^{n \times n \times d_{2}}$, where $\mathbf{b}_{ij} \in \mathbb{R}^{d_{2}}$ is the embedded feature of edge between node $\mathbf{v}_i$ and $\mathbf{v}_j$, and $d_{2}$ is the dimension of edge features.
$\mathbf{b}_{ij}$ is ``null'' if there is no edge between node $\mathbf{v}_i$ and $\mathbf{v}_j$, meaning $\mathcal{A}_{ij} = 0$.
\begin{equation}
    \mathbf{\hat{G}} = \left(\mathbf{V},\mathcal{A}, \mathbf{\hat{H}}, \mathbf{\hat{B}}\right)
    \label{eq:ESLG}
\end{equation}

$\mathbf{\hat{G}}$ is the ground truth graph, which match all the nodes and edges labels with the corresponding nodes and edges features in the graph $\mathbf{G}$.
Nodes set $\mathbf{V}$ and adjacency matrix $\mathcal{A}$ are the same in both $\mathbf{G}$ and $\mathbf{\hat{G}}$, which assumes that the graph structure remains unchanged in the transformation process. 
$\mathbf{\hat{H}} \in \{0,1,\ldots, C_1 -1\}^n$ is the nodes labels, where $\mathbf{\hat{h}}_i$ is the label of node $\mathbf{v}_i$, and $C_1$ is the number of classes for node classification tasks.
$\mathbf{\hat{B}} \in \{0,1,\ldots,C_2-1\}^{n \times n}$ is the edge labels, where $\mathbf{\hat{b}}_{ij}$ is the label of edge between node $\mathbf{v}_i$ and $\mathbf{v}_j$, and $C_2$ is the number of classes for edge classification tasks.






\subsubsection{Baseline Structure}
\label{method:end2end:baseline}
\begin{figure}
    \centering
    \includegraphics[trim={100 100 100 100},clip,width=\linewidth]{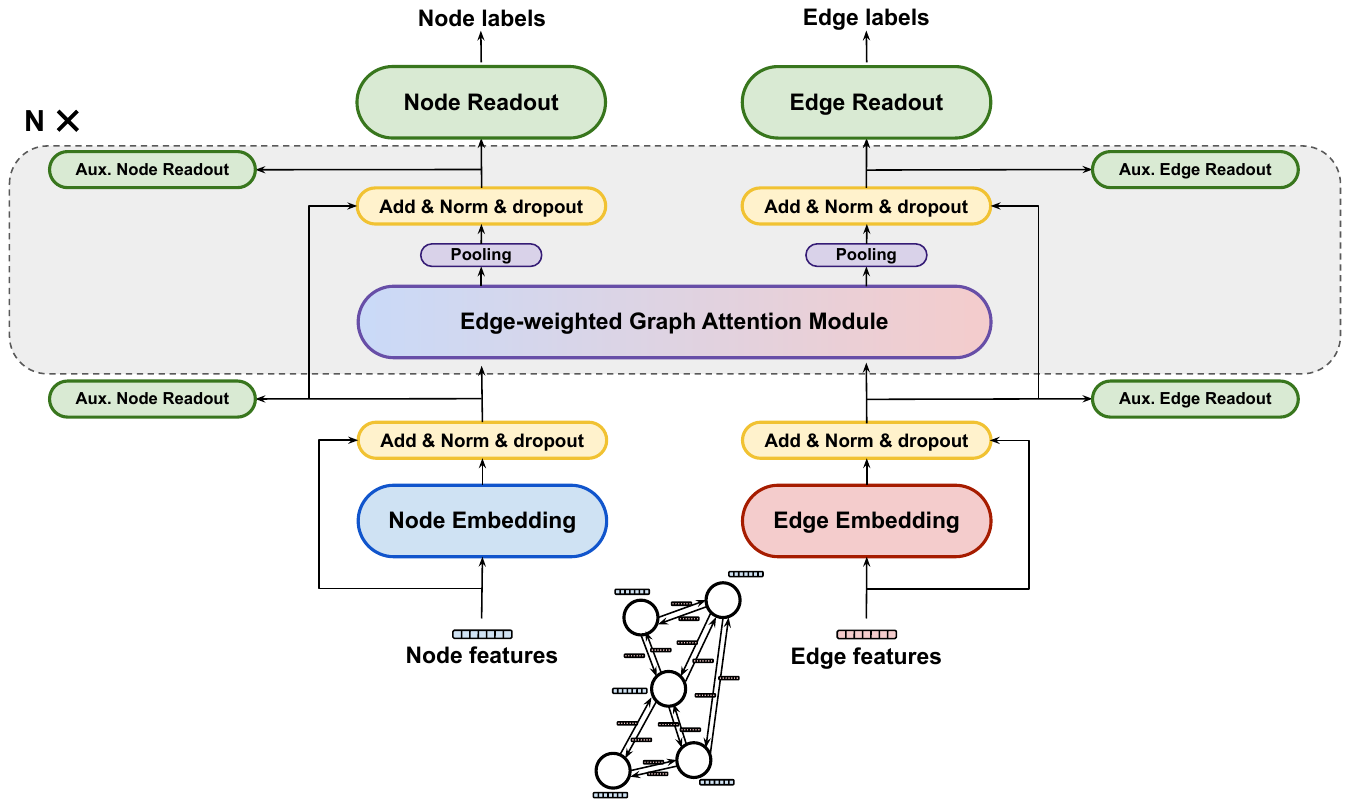}
    \caption{The baseline structure of the proposed end-to-end model.}
    \label{fig:end2end}
\end{figure}
As shown in Fig. \ref{fig:end2end}, node features and edge features are first embedded separately using the Node Embedding Module $\mathcal{E}_{node}$ and Edge Embedding Module $\mathcal{E}_{edge}$. 
\begin{equation}
    \mathbf{H}^{0} = \mathcal{E}_{node}(\mathbf{H})
\end{equation}
\begin{equation}
    \mathbf{B}^{0} = \mathcal{E}_{edge}(\mathbf{B})
\end{equation}
According to the specific characteristics of node and edge features, we could choose different neural networks for node and edge embedding modules.



The output of embedded modules $\mathbf{H}^0$ and $\mathbf{B}^{0}$ are constituted as the input of the Edge-weighted Graph Attention Module $\mathcal{G}$ for message passing and feature integration, which is the key idea of end-to-end model, and it will be detailed in section \ref{method:end2end:egat}.
\begin{equation}
    \mathbf{H}^{q}, \mathbf{B}^{q} = \mathcal{G}(\mathbf{H}^{q-1}, \mathbf{B}^{q-1}, \mathcal{A})
    \label{eq:egat}
\end{equation}
The output of $Q$-th layer $\mathbf{H}^{Q}$ and $\mathbf{B}^{Q}$ are utilized for node and edge classification tasks, respectively, with the assistance of the Node Readout $\mathcal{R}_{node}$ and Edge Readout $\mathcal{R}_{edge}$.
Both Readout networks for node features and edge features are Multi-Layer Perceptron (MLP) with the similar structure.
The last layer of the Readout networks is a softmax layer for the classification tasks.
\begin{equation}
    \mathbf{H}^\prime = \mathcal{R}_{node}(\mathbf{H}^{Q})
\end{equation}
\begin{equation}
    \mathbf{B}^\prime = \mathcal{R}_{edge}(\mathbf{B}^{Q})
\end{equation}
The output of the whole end-to-end model is the probability distribution of node labels $\mathbf{H^\prime} \in \mathbb{R}^{n\times C_1}$ and edge labels $\mathbf{B^\prime}\in \mathbb{R}^{n \times n \times C_2}$.

To enhance the model's performance, we incorporate advanced strategies like residual connections and auxiliary readout, building on the baseline structure. These will be explained in detail in the following sections.


\subsubsection{Edge-weighted Graph Attention Module}
\label{method:end2end:egat}
\begin{figure}[h!tb]
    \centering
    \begin{subfigure}[b]{0.3\textwidth}
        \includegraphics[trim={160 230 500 220},clip,width=\textwidth]{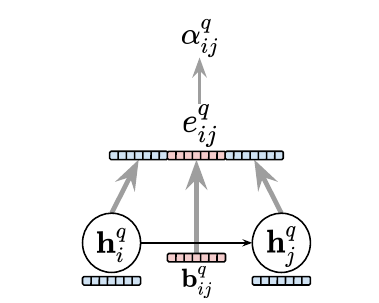}
        \caption{Attention weight.}
        \label{fig:egat_att}
    \end{subfigure}
    \begin{subfigure}[b]{0.3\textwidth}
        \includegraphics*[trim={160 230 500 220},clip,width=\linewidth]{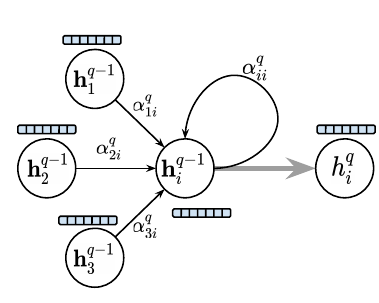}
        \caption{Node embedding update.}
        \label{fig:egat_n}
    \end{subfigure}
    \begin{subfigure}[b]{0.3\textwidth}
        \centering
        \includegraphics*[trim={160 230 500 220},clip,width=\linewidth]{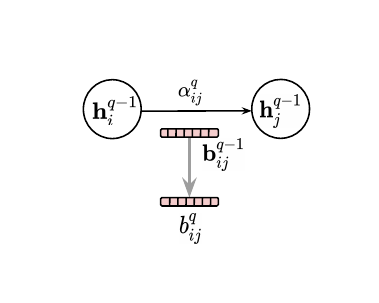}
        \caption{Edge embedding update.}
        \label{fig:egat_e}
    \end{subfigure}
    \caption{Attention Weight Computation and Message Passing, from layer $q-1$ to layer $q$ using attention $\alpha_{ij}^q$.}
    \label{fig:egat_mp}
\end{figure}

\begin{figure}[h]
    \centering
    \begin{subfigure}{0.4\textwidth}
        \centering
        \includegraphics*[trim={20 40 20 30},clip,width=\linewidth]{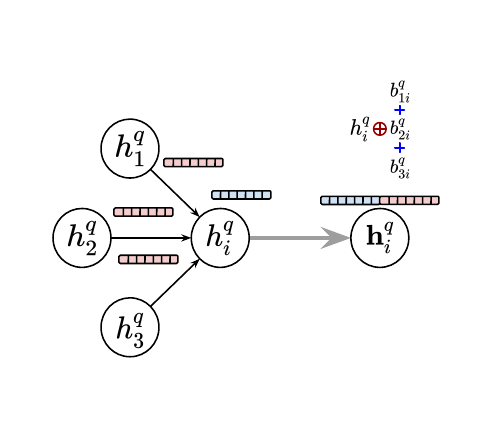}
        \caption{Node Message Concatenate}
        \label{fig:node_mc}
    \end{subfigure}
    \begin{subfigure}{0.4\textwidth}
            \centering
            \includegraphics*[trim={20 40 20 30},clip,width=\linewidth]{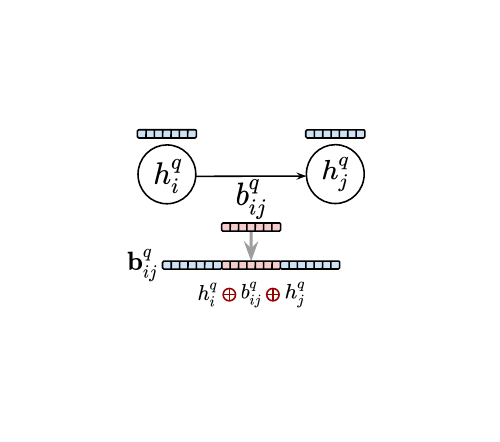}
            \caption{Edge Message Concatenate}
            \label{fig:edge_mc}
    \end{subfigure}
    \caption{Message Concatenate.}
    \label{fig:egat_mc}
\end{figure}

In each layer of the EGAT module, there are three main steps: attention weight computation in Fig.\ref{fig:egat_att}, message passing in Fig.\ref{fig:egat_n} and \ref{fig:egat_e}, and message concatenation in Fig.\ref{fig:egat_mc}.

\paragraph{Attention Weight Computation}

For layer $q$, when we calculate the attention weights, in addition to the concatenation of corresponding neighbor node features, we also concatenate the edge features together by Eq.~\ref{eq:e_ij}. 
$W_h$ and $W_b$  are learnable weights for node features and edge features, while $a$ is a learnable attention coefficient.
\begin{equation}
    e_{ij}^q = a^T\left[W_h^q\cdot \mathbf{h}_i^q \oplus  W_b^q\cdot \mathbf{b}_{ij}^q \oplus W_h^q\cdot \mathbf{h}_j^q \right]
\label{eq:e_ij}
\end{equation}

Then, with the help of softmax function, we can get the attention weights $\alpha_{ij}$ by Eq. \ref{eq:alpha_ij}. 
\begin{equation}
    \alpha_{ij}^q = \text{softmax}(e_{ij}^q) = \frac{exp(e_{ij}^q)}{\sum_{k\in \mathcal{N}_i}exp(e_{ik}^q)}
\label{eq:alpha_ij}
\end{equation}

\paragraph{Message Passing}
\label{method:end2end:egatmp}

In the $q$-th layer, the node features $h^q_i$ are updated by all the neighbor node features in last layer $\mathbf{h}^{q-1}_j$ while $j\in \mathcal{N}_i$ means all the neighbor nodes of node $i$, and also the corresponding attention weights $\alpha_{ij}^q$. 
The message passing will be achieved through the weighted summation of neighbor information, by Eq. \ref{eq:h_i_q}. 
\begin{equation}
    h_{i}^{q} = \sum_{j\in \mathcal{N} _{i}}\left(\alpha_{ij}^q \cdot W_{h}^q\cdot \mathbf{h}^{q-1}_j\right)
\label{eq:h_i_q}
\end{equation}

Likewise, the edge features $b^q_{ij}$ will also be updated by the corresponding attention weights $\alpha_{ij}$ and the edge features in last layer $b^{q-1}_{ij}$ by Eq. \ref{eq:b_ij_q}.

\begin{equation}
    b^q_{ij} = \alpha_{ij}^q \cdot W_{b}^q\cdot \mathbf{b}^{q-1}_{ij}
\label{eq:b_ij_q}
\end{equation}

In terms of interpretability, the node features compute as intermediate features, which consider the different importance of neighbor nodes and the more important node will have a larger weight in the update process according to the attention weights.
While the edge features update consider the importance of the edge itself, the attention weights indicate the importance of the edge and more important edges will be prominently emphasized during message passing.
Node and edge features $h^q_i$ and $b^q_{ij}$ updated by message passing are intermediate features in the $q$-th layer, they will be used in the following message concatenate, which is a novel concept in this paper.

\paragraph{Message Concatenate}
Nevertheless, node and edge should work as an indicator of each other in message passing, only relying on the attention scores to show the importance of the edge or neighbors is not enough.
The edge features should notice the features from the 2 nodes involved in the relationship. And the same for the node features, which should take into account the way the node is connected to its neighbor. Therefore, both updates should use both feature sources.
We proposed a further on message concatenate strategy based on above message passing by Eq.\ref{eq:h_i_q} and Eq.\ref{eq:b_ij_q}.

\begin{equation}
    \mathbf{h}^q_{i} = h_{i}^{q}\oplus \sum_{j\in \mathcal{N}_i} b_{ij}^{q}
\label{eq:H_ij_q}
\end{equation}

\begin{equation}
    \mathbf{b}^q_{ij} = h_{i}^{q}\ \oplus b_{ij}^{q} \oplus h_{j}^{q}
\label{eq:B_ij_q}
\end{equation}
The $q$-th node features $\mathbf{h}_{i}^{q}$ will be calculated by the concatenation of node features $h_{i}^{q}$ with the sum of all the connected edge features $b_{ij}^{q}$, where $j \in \mathcal{N}_i$ in Eq.\ref{eq:H_ij_q} and Fig. \ref{fig:node_mc}.
The $q$-th edge features $\mathbf{b}_{ij}^{q}$ will be calculated by the concatenation of edge features $b_{ij}^{q}$ with the connected 2 node features $h_{i}^{q}$ and $h_{j}^{q}$ in Eq.\ref{eq:B_ij_q} and Fig. \ref{fig:edge_mc}.

The mismatch between the number of parameters in the front and back layers caused by the concatenation, which will lead to an explosion in the number of parameters.
To avoid this, a pooling layer is added after the message concatenate operation for unification the dimension of the features.


\subsubsection{Advanced Optimization Techniques}
\label{method:end2end:advanced}
\paragraph{Residual Connections}
\label{method:end2end:shortcut}
Residual connections, as known as Shortcuts \cite{he2016deep}, have been proven to be effective in the training of deep neural networks, which can help to avoid the vanishing gradient problem and also enhance the deep model performance.
We applied the residual connections after Embedding module and also after each EGAT module. 
Subsequent experiments have shown that the residual connections can effectively improve the performance of baseline model both in node and edge classification tasks.
Besides, dropout layers are also added after the residual connections, which can help to avoid overfitting and improve the generalization of the model.

\paragraph{Auxiliary Readout}
\label{method:end2end:auxloss}
Inspired by GoogLeNet \cite{szegedy2015going}, auxiliary classifiers can be added to the intermediate layers of the deep neural network, which is beneficial for improving the convergence of the deep network.
In this study, we added the Readout Modules after the Embedding Module and also after every layer of Edge-weighted Graph Attention Module, which can be\ used as auxiliary classifiers for the node and edge classification tasks in every intermediate layer.
To distinguish it from the Readout of the final classifier, it is called Auxiliary Readout.
There is no significant difference between Auxiliary Readout and the final Readout, which are MLP with softmax activation function for multi-class classification.
Auxiliary Readout Modules enable every intermediate layer to conduct beneficial supervision, which help the model to provide more accurate predictions.
The involvement of Auxiliary Readout makes the original bi-objective optimization problem into a multi-objective optimization problem, multi-objective loss function will be described in the following section \ref{method:end2end:multiobj}.

\subsubsection{Multi-Objective Optimization}
\label{method:end2end:multiobj}
\paragraph{Node and Edge Classification}

The node and edge classification tasks are both multi-class classification tasks, so the common loss function is cross-entropy loss, which is widely used in multi-class classification tasks.
However, according to different tasks, more advanced loss functions can be applied to improve the performance of the model.
In math recognition, the edge classification is not balanced, so we applied the cross-entropy loss $\mathcal{L}_{n}$ for node classification, and also the focal loss $\mathcal{L}_{e}$ for edge classification task, which can help to focus on the hard samples and also the unbalanced samples.
\begin{equation}
    \mathcal{L} = \lambda_1 \mathcal{L}_{n} + (1-\lambda_1) \mathcal{L}_{e}
\end{equation}

The binary optimization loss $\mathcal{L}$ is a weighted sum of node classification loss $\mathcal{L}_{n}$ and edge classification loss $\mathcal{L}_{e}$, where $\lambda_1$ is the weight factor.

\paragraph{Auxiliary Loss}
Auxiliary loss is used to supervise the intermediate Auxiliary Readout, which can help the model to converge faster and more stable for the entire model.
We calculate the auxiliary loss $\mathcal{L}_{aux}^i$ for the $i^{th}$ Auxiliary Readout, which is the same as the node classification loss and edge classification loss, as shown in Eq. \ref{eq:aux_loss}.
\begin{equation}
    \mathcal{L}_{aux}^i = \lambda_1 \mathcal{L}_{n}^i + (1-\lambda_1) \mathcal{L}_{e}^i
\label{eq:aux_loss}
\end{equation}
The final multi-objective optimization loss $\mathcal{L}$ is a weighted sum of node classification loss, edge classification loss and also the auxiliary loss, as shown in Eq. \ref{eq:multi_loss}. 
We set the weight factor $\lambda_2$ for the auxiliary loss, and as well as the weight factor $\lambda_1$ for the node classification loss and edge classification loss, which is the same for each layer for simplicity.
$M$ is the number of Auxiliary Readout in the whole end-to-end structure.
\begin{equation}
    \mathcal{L} = \lambda_1 \mathcal{L}_{n} + (1-\lambda_1) \mathcal{L}_{e} + \sum_{i=1}^{M}\lambda_2 \mathcal{L}_{aux}^i
\label{eq:multi_loss}
\end{equation}




\subsection{Graph Modeling for HMER}
\label{method:modelisation}
\begin{figure}[h!tb]
    \centering
    \begin{subfigure}{0.3\textwidth}
        \centering
        \includegraphics*[width=\linewidth]{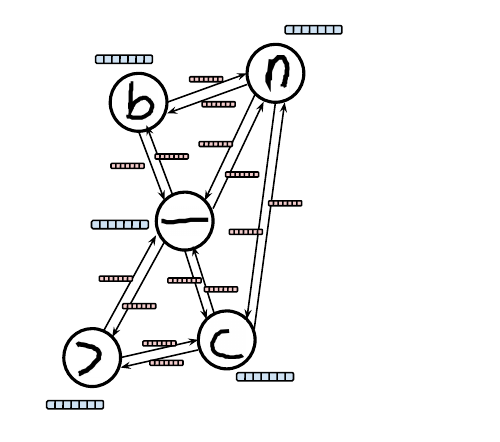}
        \caption{Local Graph Modeling.}
        \label{fig:localmodeling}
    \end{subfigure}
    \begin{subfigure}{0.3\textwidth}
            \centering
            \includegraphics*[width=\linewidth]{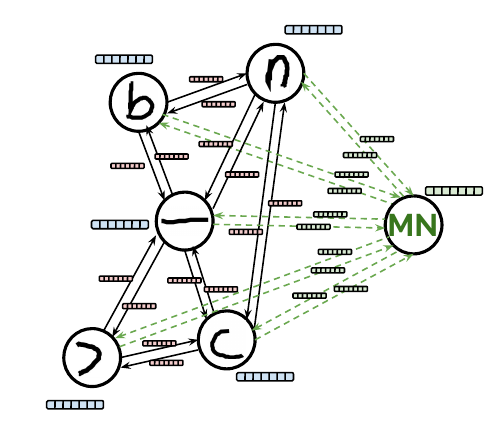}
            \caption{Global Graph Modeling with master node (MN).}
            \label{fig:globalmodeling}
    \end{subfigure}
    \caption{Illustration of the Local and Global Graph Modeling for HMER.}
    \label{fig:graphmodeling}
\end{figure}

We introduce the Graph Modeling method for HMER, which transfers the Online HME into a graph structure, and then applies the proposed graph-to-graph model for the node and edge classification tasks.
The Graph Modeling method is based on the Stroke-level Graph Modeling, each node represents a stroke in Online HME, and each edge represents the position relation between two strokes.
The fundamental Stroke-level Graph Modeling is detailed in section \ref{method:local}, which captures the local information of the Online HME.
However, it misses the global information of the whole expression.
To address this, we propose a Global Graph Modeling method in section \ref{method:global}, which is built on the Local Graph Modeling, to capture the full structure of the entrie expression.
Local and Global Graph Modeling for HMER is shown in Fig. \ref{fig:graphmodeling}.

\subsubsection{Local Graph Modeling}
\label{method:local}
Stroke-level Graph Modeling which effectively captures the local information of Online HME, we denoted this as Local Graph Modeling (LGM) $\mathbf{G}^L$, which is a similar modeling method as our previous work \cite{xie2024strokelevelgraph}.
Each stroke $\mathbf{v}_i^L \in \mathbf{V}^L$ is represented by a node in Online HME.
It also requires detailed embedding representation of node features $\mathbf{H}^L$, edge features $\mathbf{B}^L$, and the connectivity $\mathcal{A}^L$ between the nodes, without considering the global information combination of the full expression.


\paragraph{Graph Connectivity Construction}
\label{method:los}

We applied the Line-of-Sight (LOS)\cite{hu2016line} method to construct the connected edges between nodes. 
This representation of online handwritten strokes is computed using stroke convex hulls in spatial positions, which indicates the visibility between strokes. 
Based on LOS, we also add a new undirected connection from neighbors of temporal order, which increases the connectivity of the graph, and it is noted as LOS+t graph.
\begin{equation}
    \mathcal{A}^L_{ij} = \begin{cases}
        1 & \text{if $j$ is visible from $i$ or temporal neighbor of $i$,}
         \\
        0 & \text{otherwise.}
        \end{cases}
\end{equation}
Thus, the adjacency matrix $\mathcal{A}^L \in \{0,1\}^{n\times n}$ provides the connectivity relationships between nodes, where $n$ is the number of strokes in Online HME.
All the edges in graph $\mathbf{G}^L$ are bidirectional edges, that means $ \mathcal{A}^L_{ij} = \mathcal{A}^L_{ji}$.



\paragraph{Node Modeling}
\label{method:node}

Only the $x$ and $y$ coordinates of each stroke are used as node features. However, due to variations in data collection devices and user writing habits, stroke coordinates for the same symbol class can differ significantly.
To mitigate these variations, standardization is necessary. We applied a strategy to remove writing speed effects \cite{pastor2005writing} from the raw stroke coordinates, along with Gaussian normalization according to average diagonal length of all strokes of the expression.
The standardized stroke features $\mathbf{h}_i \in \mathbb{R}^{2 \times d_n}$ of node $\mathbf{v}^L_i \in \mathbf{V}^L$, where $d_n$ is the number of sampling points per stroke for node modeling.
The matrix $\mathbf{H}^L \in \mathbb{R}^{n \times 2\times d_n}$ can represent all the individual stroke features of the whole Online HME.

\paragraph{Edge Modeling}
\label{method:edge}
Edge modeling requires a strategy for modeling or extracting the relations features between strokes. The previous online documents analysis works extracted the multi-dimensional geometric features of bounding boxes of strokes, such as \cite{ye2019contextual}. 
Rather than considering the handcrafted geometric relationships between stroke bounding boxes, we aim to capture more detailed shape and bi-directional position information. 
In this study, we applied a Fuzzy Relative Positioning Template (FRPT) \cite{delaye2011fuzzy} for relations extraction.
The main idea is to calculate the radian degree between standard vectors $\overrightarrow{e^x}$ in 4 directions (Right$\rightarrow$, Left$\leftarrow$, Up$\uparrow$ and Down$\downarrow$) and the vector $\overrightarrow{OP_k}$ from the center of initial stroke $O$ to the sampling points of target stroke $P_k$, as shown in Eq. \ref{eq:frpt}.
The connections between the nodes always have 2 directions with different embeddings.
\begin{equation}
    \theta_k^x = max\left(0, 1-\frac{2}{\pi}\arccos \left(\frac{\overrightarrow{OP_k}\cdot \overrightarrow{e^x}}{\sqrt{|\overrightarrow{OP_k}|^2 + |\overrightarrow{e^x}|^2}} \right) \right)
    \label{eq:frpt}
\end{equation}
Besides the spatial relations, we also consider the distance $D_k$ between the two sampling points of the initial and target strokes.
As shown in Eq. \ref{eq:b_ij}, the edge features $\mathbf{b}_{ij} \in \mathbb{R}^{5 d_2}$ of edge $\mathbf{e}_{ij}^L \in \mathbf{E}^L$ are represented by the concatenation of the 4 directions and the distance, where $d_2$ is the number of sampling points per stroke for edge modeling.
${b}_{ij} \neq {b}_{ji}$ for the bi-directional edge, since the initial and target strokes are different.
\begin{equation}
    \mathbf{b}_{ij} = \left[\theta^\rightarrow_0, \ldots,\theta^\rightarrow_{d_2},\theta^\leftarrow_0, \ldots,\theta^\leftarrow_{d_2},\theta^\uparrow_0, \ldots,\theta^\uparrow_{d_2},\theta^\downarrow_0, \ldots,\theta^\downarrow_{d_2},D_{0}, \ldots, D_{d_2}  \right]
    \label{eq:b_ij}
\end{equation}
Thus, the matrix $\mathbf{B}^L \in \mathbb{R}^{n\times n\times 5d_2}$ represents the relationship features between strokes.

\paragraph{Sub-Expression Splitting and Masking}
\label{method:subsplit}
In real-world applications, the number of strokes in Online HME varies, with complex expressions containing many strokes and simple ones containing only a few. 
We split the Online HME into multiple sub-expressions $\mathbf{G}^{sub}$ if the stroke count exceeds $N_{max}$. 
Expressions or sub-expressions with fewer than $N_{max}$ strokes are padded with a special blank stroke, where all coordinates are set to 0.
Node features $\mathbf{H}^{sub} \in \mathbb{R}^{N_{max}\times d_1}$ and edge features $\mathbf{B}^{sub} \in \mathbb{R}^{N_{max}\times N_{max}\times d_2}$ are constructed for each sub-expression.

The sub-expression splitting is done solely based on the order of stroke input, without considering grammar rules, spatial relations, or symbolic integrity. As a result, multi-stroke symbols may be split across different sub-expressions, leading to incomplete symbols that complicate recognition.
To address this, we mask incomplete symbols within sub-expressions.
If $\exists \mathbf{v}_i \in \mathbf{V}^{sub}$, $\exists \mathbf{v}_j \notin \mathbf{V}^{sub}$ and the relation label $\mathbf{\hat{b}}_{ij}$ between $\mathbf{v}_i$ and $\mathbf{v}_j$ is ``*'', that means the symbol is split across sub-expressions.
We mask both the incomplete node features $\mathbf{h}^{sub}_i$, the relative edge features $\mathbf{b}^{sub}_{ik}$ and also their ground-truth labels $\mathbf{\hat{h}}^{sub}_i$ and $\mathbf{\hat{b}}^{sub}_{ik}$, where $\mathcal{A}_{ik}=1$.
These nodes and edges features are not involved in the training loss calculation and backward propagation, ensuring only complete symbols are used for training.
The splitting and masking strategy is applied exclusively during training to enable multi-batch processing, while the validation and testing stages use the original full graph modeling.
This approach enhances the model's robustness to local features, improving the recognition of individual symbols and their relations. However, focusing too much on local information risks overlooking the global structure of the whole expression.

\subsubsection{Global Graph Modeling}
\label{method:global}
To capture the global information of the full expression, we proposed a simple global graph modeling method by adding a master node to the local graph modeling.

\paragraph{Master Node and Edge Connection}
In terms of interpretability, deep GNNs can capture information from neighbors at increasing depths. However, relying solely on deep networks makes it challenging to effectively aggregate high-level global information.

To address this, the concept of a master node has proven effective in graph-based models for capturing global information, particularly in deep networks. Therefore, we introduced a master node $\mathbf{v}_M$ in global graph modeling, which is connected to all nodes in the original local graph modeling, as shown in Fig. \ref{fig:globalmodeling}. 
The nodes set of the global graph $\mathbf{V}^G = \mathbf{V}^L \cup \mathbf{v}_M$, and the adjacency matrix $\mathcal{A}^G \in \mathbb{R}^{(n+1)\times (n+1)}$ is calculated with Eq.\ref{eq:a_ij^G}, based on $\mathcal{A}^L$.
\begin{equation}
    \mathcal{A}^G_{ij} = \begin{cases}
        1 & \text{if $i=0$ or $j=0$ or $\mathcal{A}^L_{(i-1)(j-1)}=1$}
         \\
        0 & \text{otherwise.}
        \end{cases}
    \label{eq:a_ij^G}
\end{equation}

Through the message-passing process, the virtual master node automatically aggregates global information from all local nodes, and the information retrieved at each layer is propagated to subsequent layers.





\paragraph{Feature Initialization}
The initial features of the master node and the connected edges should be additionally taken into account.
To better integrate the shape information from all the strokes in the entire expression, rather than sub-expression split by \ref{method:subsplit}, the initial features of the master node $\mathbf{h}_M$ used as the summation of all the node features in the original modeled graph, which denotes $\mathbf{h}_M =\sum_{\mathbf{v}_i \in \mathbf{V}^L} \mathbf{h}_i$ and $\mathbf{H}^G = \left[ \mathbf{h}_M, \mathbf{h}_0, \mathbf{h}_1, \ldots, \mathbf{h}_{n} \right]^T$.
The connected edges between the master node and the local nodes are initialized as vector 0 to avoid spatial confusion, as shown in Eq. \ref{eq:b_ij^G}.
\begin{equation}
    \mathbf{B}^G_{ij} = \begin{cases}
        \mathbf{B}^L_{(i-1)(j-1)} & \text{if $i \neq 0$ and $j \neq 0$}
         \\
        0 & \text{otherwise.}
        \end{cases}
    \label{eq:b_ij^G}
\end{equation}

\subsubsection{Edited Stroke Label Graph}
\label{method:e-SLG}
\begin{figure}
    \centering
    \begin{subfigure}{0.3\textwidth}
        \centering
        \includegraphics*[trim={130 200 500 200},clip,width=\linewidth]{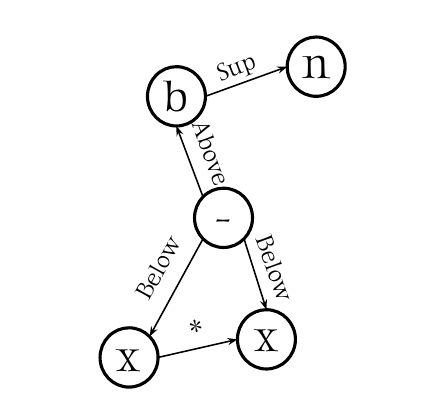}
        \caption{SLG}
        \label{fig:SLG}
    \end{subfigure}
    \begin{subfigure}{0.3\textwidth}
            \centering
            \includegraphics*[trim={130 200 500 200},clip,width=\linewidth]{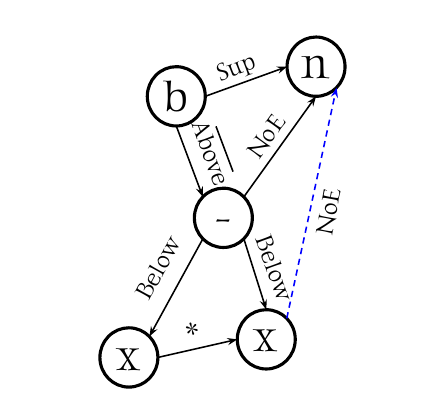}
            \caption{ESLG}
            \label{fig:ESLG}
    \end{subfigure}
    \caption{From Stroke Label Graph (SLG) to Edited-SLG (ESLG) for expression $\frac{b^n}{x}$ written with 5 strokes.
    ``Below'', ``Above'', ``Sup'' are the positional relations, ``\barre{Above}'' is the opposite positional relation of ``Above''.
    While ``*'' means belonging relation and ``NoE'' means no relation.}    
    \label{fig:SLG-ESLG}
\end{figure}

The results of HMER can be represented as different formats, such as MathML tree, LaTeX string, and also the stroke-level graph called Stroke Label Graph (SLG) proposed by \cite{mouchere2012icfhr}, which is a directed graph structure, where the nodes represent the strokes belonging to which symbol and the edges represent the relations between the strokes, Fig \ref{fig:SLG} shows an example of Stroke Label Graph.
Based on SLG, we make some simple edge modifications to match the nodes and edges of SLG with the modelised stroke-level OHME graph, which is denoted as Edited Stroke Label Graph (ESLG), $\mathbf{\hat{G}}$ in Eq. \ref{eq:ESLG}.
The relations in mathematical expression are all directional according to the relative position.
For example, ``$b$'' is above ``$-$'' in the expression $\frac{b^n}{x}$, so the edge ``above'' should point from ``$-$'' to ``$b$''.
In LOS+t graph, all the edges are bidirectional, the high dimension features of edge ``$-$'' to ``$b$'' should have the information means ``Above'', while the edge ``$b$'' to ``$-$'' should have the information means opposite of ``Above'', we noted as ``\barre{Above}''. 
However, labeling both directions of bidirectional edges can lead to confusion in the final decision process. To resolve this, we retain only one directed edge for each pair of connected strokes, following the writing order. This means each edge is directed from the earlier stroke to the later stroke, as calculated in Eq. \ref{eq:hat_a_ij}.
\begin{equation}
    \mathcal{\hat{A}}_{ij} = \begin{cases}
        1 & \text{if $j > i$ and $\mathcal{A}_{ij}=1$}
         \\
        0 & \text{otherwise.}
        \end{cases}
    \label{eq:hat_a_ij}
\end{equation}
This strategy maintains the structure of the SLG and disregards syntactic rules, but avoids confusion of directions through temporal ordering.
The edge between ``$-$'' and ``$b$'' is kept as ``\barre{Above}'' from ``$b$'' to ``$-$'' according to the writing order.
In addition, the ``*'' relation, which indicates that two strokes belong to the same symbol, and the ``NoE'' (No Edge) relation, which applies to edges present in the LOS+t graph but absent in the SLG, signifying no relation between the two strokes. An example of the edited SLG is illustrated in Fig. \ref{fig:ESLG}.
Thus, ESLG $\mathbf{\hat{G}}$ has the same node set $\mathbf{V}$ as the corresponding modeling graph $\mathbf{G}$, while $\mathbf{\hat{H}} \in \mathbb{R}^ {n \times C_{1}}$ is the node labels of entire Online HME belonging to $C_{1} = C_{n}$ classes of symbols, and $\mathbf{\hat{B}} \in \mathbb{R}^{n \times n \times C_2}$ is the edge labels of entire Online HME belonging to $C_2 = 2C_e +2$ classes of relations including $C_e$ positional relations, $C_e$ opposite positional relations, ``*'' and ``NoE''.



\section{Experiments}
\label{exp}
In this section, we present the experimental results of the proposed end-to-end framework with Edge-weighted Graph Attention Network for HMER problem, which is noted as LGM-EGAT for Local Graph Modeling and GGM-EGAT for Global Graph Modeling.
\subsection{Experiments Setup}
\subsubsection{Datasets}
\label{exp:datasets}
The proposed model is trained and evaluated on the newest Competition on Recognition of Online Handwritten Mathematical Expression dataset (CROHME 2023) \cite{xie2023icdar}.
The CROHME 2023 provides large-scale on-line OHME data in InkML \footnote{{http://www.w3.org/2003/InkML}} format, and provides the ground truth with expressions-level, symbol-level, and stroke-level annotations in \LaTeX, MathML and SLG format.
Using the ``LG'' format SLG files, 101 symbol classes ($C_n=101$) and 6 relation classes (``Right'', ``Sup'', ``Sub'', ``Above'', ``Below'', ``Inside'')—where $C_e=6$—are labeled in the SLG.

\subsubsection{Implementation Details}
\label{exp:impdetails}
Our proposed model is implemented by PyTorch-Lightning framework, and trained on a single NVIDIA GeForce RTX 2080 Ti GPU with 12GB memory.
All the hyperparameters are selected by fine-tuning with optuna framework, initial learning rate $\eta = 0.00027$, the number of sub-expressions in batch training $B=32$, maximum number of strokes in sub-expression $N_{max}=16$, bi-objective optimization loss weight $\lambda_1=0.5$, auxiliary loss weight $\lambda_2=0.3$, focal loss focusing parameter $\gamma=1.5$ and dropout rate $p=0.1$.
The model is trained with Adam optimizer, and the learning rate is decayed by a factor of 0.1 if the validation loss does not decrease for 20 epochs, the training process is stopped after 200 epochs.
The number of sampling points for node modeling $d_1=150$, the number of sampling points for edge modeling $d_2=10$, and the model configuration is shown in Table \ref{tab:model_config}.
XceptionTime \cite{rahimian2020xceptiontime} is utilized for node embedding, which is a 1D convolutional neural network designed for time series data. While a simple Multi-Layer Perceptron (MLP) is used for edge embedding.
\begin{table}[h]
    \centering
    \begin{tabular}{ccc}
        \hline
        \textbf{Module Name}  & \textbf{Details} & \textbf{Module nb.}\\
        \hline
        Node Embedding  & $\mathcal{E}_n = \text{XceptionTime}(2,150)$& 1 \\
        Edge Embedding & $\mathcal{E}_e =\text{MLP}(50,384,512)$  & 1\\
        EGAT Layer & $\mathcal{G}_{node}=(512,512), \mathcal{G}_{edge}=(512,512)$ & 5  \\
        Node Readout  & $\mathcal{R}_n = \text{MLP}(1024,384,101)$& 1 \\
        Edge Readout & $\mathcal{R}_e = \text{MLP}(1024,384,14)$ & 1 \\
        Aux. Node Readout & $\mathcal{R}_n^{aux} = \text{MLP}(1024,384,101)$ & 5 \\
        Aux. Edge Readout  & $\mathcal{R}_e^{aux} = \text{MLP}(1024,384,14)$& 5 \\
        \hline
    \end{tabular}
    \caption{End-to-end Model Configuration.}
    \label{tab:model_config}
\end{table}


\subsection{Results}
\label{exp:results}
In this section, we present enriched ablation studies for the proposed end-to-end model, and compare the performance of the proposed model with the state-of-the-art methods.
\subsubsection{Ablation Study}
\label{exp:ablation}
We conducted a series of experiments to evaluate the performance of the proposed HMER system with different end-to-end model structures, graph modeling strategies and graph connectivity construction methods.
All of the experiments are evaluated on the validation set of CROHME 2023. 
The metrics are the accuracy of classification (Clf. Acc.) for both node and edge in primitive levels. 
The correct symbol segmentation rate (Seg.), the correct symbol segmentation and symbol classification rate (Sym.), the correct relation classification rate (Rel.), the correct expression recognition rate (Exp.) and the correct structure recognition rate (Struc.) are accuracy in expression level provided by LgEval\cite{mouchere2014icfhr}.

As shown in Table \ref{tab:ab1}, in order to compare the effectiveness of the proposed end-to-end model, we applied the global graph modeling strategy and ``LOS+t'' as the graph connectivity construction method.
The baseline model is the end-to-end model only with node and edge embedding network, EGAT layers, and node and edge readout networks, which was proposed in our previous work \cite{xie2024strokelevelgraph}.
Based on this baseline model, we added the residual connection (BL+Shortcut), auxiliary loss (BL+Aux. Loss), and also the message concatenate (BL+Mes. Cat.) we proposed in the previous sections.
The proposed model is the baseline model with all of advanced strategies.
According to the results, the message concatenate strategy significantly improves edge classification performance, while the auxiliary loss and residual connection enhance both node and edge classification tasks.
The proposed model achieves the best accuracy in both node and edge classification tasks, which demonstrates the effectiveness of the advanced strategies.
\begin{table}[h!tp]
    \centering
    \scalebox{0.8}{
    \begin{tabular}{c|cc|ccccc}
    \hline
    &\multicolumn{2}{c|}{\textbf{Clf. Acc.}}& \multicolumn{5}{c}{\textbf{Exp. level}}\\ 
    \hline
    \textbf{End2end} & \textbf{Node} &\textbf{Edge}  & \textbf{Seg.} & \textbf{Sym.}& \textbf{Rel.} & \textbf{Exp.} & \textbf{Stru.}\\
    \hline
    Baseline&92.42& 95.60 & 97.54 & 90.88 & 90.60 & 45.30 & 71.50 \\
    BL + Shortcut&93.21& 95.89 & 97.56 & 91.63 & 91.29& 48.18 & 72.91\\
    BL + Aux. Loss& 93.05 & 95.79 & 97.77 & 91.66 & 91.25 & 47.83 &72.33\\
    BL + Mes. Cat. & 91.30& 96.80 & 98.13 & 89.60 & 92.20 & 42.95 & 75.56\\
    Proposed& \textbf{94.40} & \textbf{97.50} & \textbf{98.39} &  \textbf{93.34} &\textbf{93.69} & \textbf{55.88} &\textbf{80.73}\\
    \hline
    \end{tabular}}
    \caption{Ablation Study for end-to-end model and graph modelisation.
    All of the experiments are evaluated on the validation set of CROHME 2023. 
    The metrics are the accuracy of classification (Clf. Acc.) for both node and edge in primitive level. 
    The correct symbol segmentation rate (Seg.), the correct symbol segmentation and symbol classification rate (Sym.), the correct relation classification rate (Rel.), the correct expression recognition rate (Exp.) and the correct structure recognition rate (Struc.) are accuracy at expression level.}
    \label{tab:ab1}
\end{table}

In Table \ref{tab:ab2}, we applied both local and global graph modeling strategies, and compared the performance of the proposed model with the baseline model.
According to node and edge classification accuracy, the global graph modeling strategy slightly improves both node and edge classification performance. 
Although the segmentation accuracy, symbol recognition accuracy, and relation recognition accuracy have not improved significantly even lower, the expression recognition and structure recognition accuracy in expression level have an upgrade.
It is implied that the global graph modeling strategy can enhance the global information capture and improve the recognition of the entire expression, which is beneficial for the expression recognition and structure recognition tasks.

\begin{table}[h!tp]
    \centering
    \scalebox{0.8}{
    \begin{tabular}{c|c|cc|ccccc}
    \hline
    \multicolumn{2}{c|}{}&\multicolumn{2}{c|}{\textbf{Clf. Acc.}}& \multicolumn{5}{c}{\textbf{Exp. level}}\\ 
    \hline
    \textbf{End2end} &\textbf{G.M} & \textbf{Node} &\textbf{Edge}  & \textbf{Seg.} & \textbf{Sym.}& \textbf{Rel.} & \textbf{Exp.} & \textbf{Stru.}\\
    \hline
    BL&L.& 92.12 & 94.78 & 97.56 & 90.20 & 90.55& 44.89 & 71.62 \\
    BL&G.& 92.42 & 95.60 & 97.54 & 90.88 & 90.60 & 45.30 & 71.50 \\
    Proposed&L.& 94.16 & 97.48 & \textbf{98.44} & \textbf{93.45} & \textbf{93.79} & 55.23 & 79.85\\
    Proposed& \textbf{G.} &\textbf{94.40} & \textbf{97.50} & 98.39&  93.34 &93.69 & \textbf{55.88} &\textbf{80.73}\\
    \hline
    \end{tabular}}
    \caption{Ablation Study for Graph Modeling Strategies. ``BL'' is Baseline end2end model proposed in \cite{xie2024strokelevelgraph}, ``Proposed'' is the proposed model in this paper with message concatenate and advanced strategies. ``G.M'' is the graph modeling strategy, while ``L.'' is local graph modeling, ``G.'' is global graph modeling. All of the evaluation metrics are the same as Table \ref{tab:ab1}.}
    \label{tab:ab2}
\end{table}

In Table \ref{tab:ab3}, we explored the effectiveness of the graph connectivity connection.  
Compared with the Full Connected (FC) graph, the proposed ``LOS+t'' strategy significantly improves the performance in all the evaluation metrics, especially with poorer information fusion model like the baseline model.
That means the proposed ``LOS+t'' strategy can help the end-to-end model learn more specific and accurate graph structure information, which is beneficial for the expression and structure recognition tasks.
Meanwhile, the proposed end2end model with strong information fusion ability is able to learn the complex graph connectivity connection relation, and could achieve better performance with the help of  ``LOS+t'' strategy.
\begin{table}[h!tp]
    \centering
    \scalebox{0.8}{
        \begin{tabular}{c|c|cc|ccccc}
    \hline
    \multicolumn{2}{c|}{}&\multicolumn{2}{c|}{\textbf{Clf. Acc.}}& \multicolumn{5}{c}{\textbf{Exp. level}}\\ 
    \hline
    \textbf{End2end} &\textbf{G.C.C.} & \textbf{Node} &\textbf{Edge}  & \textbf{Seg.} & \textbf{Sym.}& \textbf{Rel.} & \textbf{Exp.} & \textbf{Stru.}\\
    \hline
    BL&FC& 82.11 & 91.16 & 96.30 & 79.07 & 86.27& 22.50 & 58.34 \\
    BL&LOS+t& 92.42 & 95.60 & 97.54 & 90.88 & 90.60 & 45.30 & 71.50 \\
    Proposed&FC& 93.88 & 97.12 & 98.05& 92.84 & 92.31 & 52.00 & 76.62\\
    Proposed& LOS+t&\textbf{94.40} & \textbf{97.50} & \textbf{98.39} &  \textbf{93.34} &\textbf{93.69} & \textbf{55.88} &\textbf{80.73}\\
    \hline
    \end{tabular}}
    \caption{Ablation Study for Graph Connectivity Connection. ``BL'' is Baseline end2end model proposed in \cite{xie2024strokelevelgraph}, ``Proposed'' is the proposed model in this paper with message concatenate and advanced strategies. ``G.C.C.'' is the graph connectivity connection, while ``FC'' is Full Connected, ``LOS+t'' is the proposed strategy. All the evaluation metrics are the same as Table \ref{tab:ab1}.}
    \label{tab:ab3}
\end{table}

\subsubsection{Comparison with State-of-the-Art}
\label{exp:comparison}

In this section, we compare our proposed LGM-EGAT (Local Graph Modeling with Edge-weight Graph Attention Mechanism) and GGM-EGAT (Global Graph Modeling with Edge-weight Graph Attention Mechanism) systems with state-of-the-art methods. 
Since there are only a few methods focusing on online handwritten mathematical expression recognition, we also take some offline methods into consideration for comparison.
For fair comparison, we fine-tuned and evaluated our proposed models on the CROHME 2016 and CROHME 2019 data sets.
The results are presented in Table \ref{tab:sota1}. The proposed LGM-EGAT and GGM-EGAT systems outperform the state-of-the-art methods on both CROHME 2016 and CROHME 2019 datasets, including traditional encoder-decoder methods \cite{zhang2017watch, zhang2018track}, tree-based encoder-decoder methods\cite{zhang2017tree, truong2021learning}, and also encoded-decoder methods with graph neural networks \cite{wu2021graph,tang2024offline}.
All listed models are trained and evaluated on the same dataset without any additional data augmentation or pre-training strategies.
\begin{table}[h!tp]
    \centering
    \scalebox{0.8}{
    \begin{tabular}{c|c|c|c}
        \hline
        \textbf{Systems}  
        & \begin{tabular}[c]{@{}c@{}}\textbf{Data} \\ \textbf{Type}\end{tabular}   
        & \begin{tabular}[c]{@{}c@{}}\textbf{CROHME}\\\textbf{2016}\end{tabular}
        & \begin{tabular}[c]{@{}c@{}}\textbf{CROHME}\\\textbf{2019}\end{tabular} \\
        \hline
        WAP\cite{zhang2017watch} & Off. & 44.45 & - \\
        GETD\cite{tang2024offline} & Off. & 55.27 & 54.13 \\
        TAP\cite{zhang2018track} & On. & 44.80 & - \\
        Tree-construction\cite{truong2021learning} &  On. & 41.76 & - \\
        Tree-BiLSTM\cite{zhang2017tree} & On.  & 27.03 &- \\
        G2G\cite{wu2021graph} & On. & 52.05 & - \\
        \textbf{LGM-EGAT} & On. & \textbf{56.41} & \textbf{58.22} \\
        \textbf{GGM-EGAT} & On. & \textbf{56.67} & \textbf{60.72} \\
        \hline
    \end{tabular}}
    \caption{Comparison with State-of-the-Art. ``On.'' is model with online data, ``Off.'' is model with offline data. ``-'' means the result is not available from the original paper. The metric for both CROHME 2016 and CROHME 2019 is the expression recognition accuracy.}
    \label{tab:sota1}
\end{table}

In Table \ref{tab:sota2}, we compare our proposed LGM-EGAT and GGM-EGAT systems with the top-ranked methods in the CROHME 2023 competition.
The proposed models are trained without any additional data augmentation or pre-training strategies, but it still shows the potential especially in structure correctness recognition.

\begin{table}[h!tp]
    \centering
    \scalebox{0.8}{
    \begin{tabular}{c|cccc|c|cccc}
        \hline
        \textbf{2019} & Exp. & $\le 1$& $\le 2$  & Stru. &\textbf{2023} & Exp. & $\le 1$& $\le 2$ &Stru.\\
        \hline
        iFLYTEK*\ddag & 80.73 & 88.99 & 90.74 & 91.49 &Sunia\dag & 82.34 & 90.26 & 92.47 & 92.41\\
        Samsung\dag & 79.82 & 87.82 & 89.15 & 89.32 &YP\_OCR* & 72.55 & 83.57 & 86.22 & 86.60\\
        MyScript\dag\ddag & 79.15 & 86.82 & 89.82 & 90.66 &TUAT & 41.10 & 54.52 & 60.04 & 56.85\\
        PAL-v2*\ddag & 62.55 & 74.98 & 78.40 & 79.15 &DPRL & 38.19 & 53.39 & 58.39 & 59.98\\
        MathType\dag & 60.13 & 74.40 & 78.57 & 79.15 &-&-&-&-&-\\
        TUAT & 39.95 & 52.21 & 56.54 & 58.22&-&-&-&-&-\\
        \hline
        LGM-EGAT & 58.22 & 70.89 & 75.48 & 83.07& LGM-EGAT & 53.04 & 66.26 & 72.22 & 78.39\\
        GGM-EGAT & 60.72 & 71.14 & 76.73 & 83.74 &GGM-EGAT & 55.30 & 68.43 & 72.91 & 79.13\\
        \hline
    \end{tabular}}
    \caption{Comparison with CROHME Competition 2019 and 2023. All the results are according to CROHME 2019 and 2023. ``Exp.'' in expression level is the expression recognition correct accuracy, while ``$\le 1$'' and ``$\le 2$ '' are the expression recognition accuracy with less than 1 and 2 errors, respectively. The ``Stru.'' is the structure recognition accuracy.``*'' denotes an ensemble of several differently initialized recognition models. ``\dag'' denotes the model is trained with additional data. ``\ddag'' denotes using a mathematical Language Model and additional \LaTeX{} sequences.}
    \label{tab:sota2}
\end{table}

\subsection{Analysis}
\label{analysis}
\subsubsection{Attention Visualization}
\label{analysis:attention}
We save the attention scores of the last EGAT layer, which usually represent the importance of the neighbors of one node, and also the importance of the connected edges.
\begin{figure}[h]
    \centering

    \begin{subfigure}{0.4\textwidth}
        \centering
        	\tikzset{stroke/.style={black, line width=0.1mm}}

	\begin{tikzpicture}
\draw [stroke] (0.0,0.64) -- (0.07,0.79) -- (0.09,0.79) -- (0.1,0.79) -- (0.11,0.77) -- (0.13,0.76) -- (0.14,0.73) -- (0.14,0.72) -- (0.14,0.69) -- (0.13,0.66) -- (0.13,0.63) -- (0.13,0.6) -- (0.11,0.59) -- (0.11,0.56) -- (0.11,0.53) -- (0.1,0.52) -- (0.1,0.49) -- (0.1,0.47) -- (0.09,0.46) -- (0.09,0.44) -- (0.07,0.44) ;
\draw [stroke] (0.34,0.9) -- (0.26,0.79) -- (0.24,0.77) -- (0.21,0.76) -- (0.2,0.73) -- (0.17,0.7) -- (0.16,0.67) -- (0.14,0.64) -- (0.14,0.62) -- (0.14,0.6) -- (0.14,0.59) -- (0.14,0.57) -- (0.16,0.56) -- (0.17,0.53) -- (0.2,0.52) -- (0.23,0.5) -- (0.26,0.5) -- (0.29,0.49) -- (0.32,0.49) -- (0.33,0.49) -- (0.36,0.5) -- (0.37,0.52) -- (0.4,0.53) -- (0.42,0.56) -- (0.46,0.57) -- (0.46,0.59) ;
\draw [stroke] (0.82,0.59) -- (0.99,0.57) -- (1.02,0.57) -- (1.05,0.59) -- (1.09,0.59) -- (1.12,0.59) -- (1.15,0.59) -- (1.17,0.59) -- (1.2,0.59) -- (1.22,0.59) -- (1.23,0.59) -- (1.23,0.6) ;
\draw [stroke] (0.9,0.76) -- (1.0,0.77) -- (1.03,0.77) -- (1.06,0.77) -- (1.09,0.77) -- (1.09,0.79) -- (1.1,0.79) -- (1.12,0.79) ;
\draw [stroke] (1.6,0.64) -- (1.75,0.64) -- (1.79,0.64) -- (1.83,0.64) -- (1.89,0.64) -- (1.96,0.64) -- (2.03,0.64) -- (2.09,0.64) -- (2.18,0.66) -- (2.26,0.66) -- (2.33,0.66) -- (2.43,0.67) -- (2.52,0.67) -- (2.61,0.67) -- (2.69,0.69) -- (2.79,0.7) -- (2.88,0.7) -- (2.95,0.72) -- (3.02,0.72) -- (3.08,0.72) -- (3.14,0.72) -- (3.16,0.72) -- (3.19,0.72) -- (3.21,0.72) -- (3.22,0.73) -- (3.24,0.73) -- (3.24,0.72) ;
\draw [stroke] (1.65,1.4) -- (1.62,1.35) -- (1.62,1.32) -- (1.62,1.29) -- (1.62,1.26) -- (1.62,1.23) -- (1.62,1.2) -- (1.62,1.17) -- (1.62,1.16) -- (1.62,1.15) -- (1.62,1.13) -- (1.62,1.12) -- (1.63,1.12) -- (1.65,1.1) -- (1.68,1.12) -- (1.69,1.12) -- (1.72,1.12) -- (1.75,1.12) -- (1.76,1.13) -- (1.79,1.13) -- (1.8,1.13) -- (1.82,1.13) -- (1.83,1.13) -- (1.83,1.15) ;
\draw [stroke] (1.86,1.26) -- (1.85,1.2) -- (1.85,1.17) -- (1.85,1.15) -- (1.85,1.13) -- (1.83,1.1) -- (1.85,1.07) -- (1.85,1.06) -- (1.85,1.05) -- (1.85,1.03) -- (1.85,1.02) ;
\draw [stroke] (2.05,1.49) -- (2.06,1.42) -- (2.06,1.39) -- (2.08,1.36) -- (2.08,1.35) -- (2.09,1.32) -- (2.09,1.29) -- (2.11,1.27) -- (2.11,1.25) -- (2.11,1.23) -- (2.11,1.22) -- (2.11,1.2) -- (2.11,1.19) ;
\draw [stroke] (2.15,1.52) -- (2.16,1.45) -- (2.16,1.42) -- (2.16,1.39) -- (2.16,1.37) -- (2.18,1.35) -- (2.18,1.33) -- (2.18,1.3) -- (2.18,1.27) -- (2.18,1.26) -- (2.18,1.25) -- (2.18,1.23) -- (2.18,1.22) -- (2.18,1.2) -- (2.18,1.22) ;
\draw [stroke] (1.96,1.42) -- (1.88,1.5) -- (1.86,1.52) -- (1.86,1.53) -- (1.85,1.53) -- (1.85,1.52) -- (1.86,1.52) -- (1.88,1.52) -- (1.89,1.52) -- (1.92,1.52) -- (1.95,1.52) -- (1.98,1.52) -- (2.0,1.52) -- (2.03,1.53) -- (2.06,1.53) -- (2.09,1.53) -- (2.11,1.55) -- (2.13,1.55) -- (2.15,1.55) -- (2.16,1.55) -- (2.18,1.56) -- (2.19,1.56) -- (2.21,1.56) -- (2.22,1.56) ;
\draw [stroke] (2.31,1.25) -- (2.38,1.32) -- (2.39,1.33) -- (2.39,1.36) -- (2.41,1.37) -- (2.41,1.39) -- (2.41,1.4) -- (2.41,1.42) -- (2.42,1.4) -- (2.42,1.39) -- (2.42,1.37) -- (2.42,1.35) -- (2.42,1.33) -- (2.43,1.32) -- (2.43,1.3) -- (2.45,1.29) -- (2.45,1.27) -- (2.45,1.26) -- (2.46,1.26) -- (2.46,1.25) -- (2.48,1.25) -- (2.49,1.25) -- (2.51,1.26) -- (2.51,1.27) ;
\draw [stroke] (2.38,1.56) ;
\draw [stroke] (2.11,0.43) -- (2.18,0.47) -- (2.21,0.47) -- (2.23,0.49) -- (2.26,0.5) -- (2.29,0.5) -- (2.31,0.5) -- (2.32,0.5) -- (2.33,0.49) -- (2.35,0.49) -- (2.35,0.47) -- (2.35,0.46) -- (2.35,0.44) -- (2.33,0.42) -- (2.32,0.4) -- (2.31,0.37) -- (2.29,0.34) -- (2.26,0.33) -- (2.25,0.32) -- (2.23,0.3) -- (2.22,0.3) -- (2.21,0.29) -- (2.19,0.29) -- (2.21,0.29) -- (2.22,0.3) -- (2.23,0.3) -- (2.25,0.32) -- (2.28,0.32) -- (2.31,0.32) -- (2.33,0.33) -- (2.36,0.33) -- (2.39,0.32) -- (2.41,0.32) -- (2.43,0.32) -- (2.45,0.3) -- (2.45,0.29) -- (2.46,0.29) -- (2.46,0.27) -- (2.46,0.24) -- (2.46,0.23) -- (2.45,0.2) -- (2.43,0.19) -- (2.42,0.16) -- (2.38,0.13) -- (2.36,0.1) -- (2.32,0.09) -- (2.28,0.06) -- (2.25,0.03) -- (2.22,0.01) -- (2.19,0.01) -- (2.18,0.0) -- (2.16,0.0) ;
\draw [stroke] (3.45,0.76) -- (3.55,0.74) -- (3.58,0.74) -- (3.61,0.74) -- (3.64,0.76) -- (3.68,0.76) -- (3.72,0.76) -- (3.78,0.77) -- (3.82,0.77) -- (3.87,0.77) -- (3.89,0.77) -- (3.92,0.77) -- (3.94,0.77) -- (3.95,0.77) ;
\draw [stroke] (3.54,0.95) -- (3.57,0.86) -- (3.57,0.84) -- (3.58,0.82) -- (3.58,0.79) -- (3.59,0.77) -- (3.59,0.74) -- (3.61,0.72) -- (3.61,0.69) -- (3.62,0.67) -- (3.62,0.66) -- (3.64,0.63) -- (3.64,0.62) -- (3.64,0.6) -- (3.65,0.6) -- (3.65,0.59) -- (3.64,0.59) ;
\draw [stroke] (4.57,1.16) -- (4.45,1.16) -- (4.44,1.17) -- (4.41,1.17) -- (4.4,1.19) -- (4.38,1.19) -- (4.37,1.2) -- (4.35,1.22) -- (4.35,1.23) -- (4.35,1.25) -- (4.35,1.26) -- (4.35,1.27) -- (4.37,1.29) -- (4.38,1.29) -- (4.4,1.3) -- (4.41,1.3) -- (4.42,1.3) -- (4.42,1.29) -- (4.44,1.27) -- (4.44,1.26) -- (4.44,1.23) -- (4.44,1.2) -- (4.44,1.19) -- (4.42,1.15) -- (4.41,1.1) -- (4.38,1.07) -- (4.35,1.03) -- (4.34,1.0) -- (4.31,0.97) -- (4.3,0.96) -- (4.28,0.93) -- (4.27,0.93) -- (4.25,0.93) -- (4.24,0.95) -- (4.24,0.96) -- (4.24,0.97) -- (4.24,0.99) -- (4.25,1.0) -- (4.27,1.0) -- (4.28,1.0) -- (4.3,0.99) -- (4.31,0.99) -- (4.32,0.97) -- (4.35,0.97) -- (4.38,0.97) -- (4.41,0.97) -- (4.45,0.97) -- (4.48,0.97) -- (4.51,0.97) -- (4.54,0.97) -- (4.55,0.97) -- (4.58,0.99) ;
\draw [stroke] (4.98,1.4) -- (4.98,1.27) -- (4.98,1.25) -- (5.0,1.22) -- (5.01,1.19) -- (5.01,1.16) -- (5.01,1.13) -- (5.03,1.1) -- (5.03,1.07) -- (5.04,1.06) -- (5.04,1.05) -- (5.04,1.03) -- (5.04,1.02) -- (5.05,1.0) ;
\draw [stroke] (5.11,1.4) -- (5.08,1.32) -- (5.1,1.3) -- (5.1,1.29) -- (5.11,1.26) -- (5.11,1.23) -- (5.13,1.2) -- (5.13,1.19) -- (5.13,1.16) -- (5.14,1.13) -- (5.14,1.12) -- (5.14,1.1) -- (5.16,1.07) -- (5.16,1.06) -- (5.16,1.05) ;
\draw [stroke] (4.9,1.27) -- (4.83,1.36) -- (4.83,1.37) -- (4.81,1.37) -- (4.81,1.39) -- (4.81,1.37) -- (4.83,1.37) -- (4.84,1.37) -- (4.85,1.37) -- (4.87,1.37) -- (4.88,1.39) -- (4.9,1.39) -- (4.93,1.4) -- (4.95,1.4) -- (4.98,1.4) -- (5.01,1.4) -- (5.05,1.42) -- (5.08,1.42) -- (5.11,1.42) -- (5.13,1.42) -- (5.16,1.42) -- (5.17,1.42) -- (5.18,1.42) -- (5.2,1.43) -- (5.23,1.43) -- (5.24,1.43) -- (5.26,1.43) -- (5.27,1.43) -- (5.27,1.45) -- (5.27,1.43) ;
\draw [stroke] (5.4,1.25) -- (5.37,1.19) -- (5.37,1.17) -- (5.37,1.15) -- (5.38,1.13) -- (5.4,1.12) -- (5.41,1.09) -- (5.41,1.07) -- (5.43,1.07) -- (5.44,1.06) -- (5.46,1.06) -- (5.47,1.07) -- (5.5,1.09) -- (5.51,1.12) -- (5.56,1.15) ;
\draw [stroke] (5.41,1.4) ;
\draw [stroke] (5.64,1.32) -- (5.67,1.23) -- (5.68,1.22) -- (5.7,1.2) -- (5.7,1.19) -- (5.71,1.19) -- (5.71,1.17) -- (5.73,1.17) -- (5.73,1.16) -- (5.73,1.17) -- (5.73,1.19) -- (5.73,1.2) -- (5.73,1.23) -- (5.74,1.25) -- (5.74,1.27) -- (5.76,1.29) -- (5.76,1.3) -- (5.77,1.3) -- (5.77,1.32) -- (5.79,1.32) -- (5.79,1.3) -- (5.8,1.3) -- (5.8,1.29) -- (5.8,1.27) -- (5.81,1.25) -- (5.81,1.23) -- (5.83,1.22) -- (5.83,1.2) -- (5.84,1.19) -- (5.84,1.17) -- (5.86,1.17) -- (5.87,1.16) -- (5.89,1.16) -- (5.9,1.16) -- (5.91,1.16) -- (5.93,1.16) -- (5.94,1.17) -- (5.96,1.17) -- (5.97,1.19) -- (5.99,1.19) -- (6.0,1.2) ;

		   \end{tikzpicture}
        \caption{Ex.1 $x = \frac{4\pi_i}{3} + 3 \pi _i n$}
        \label{fig:attention:ex1}
    \end{subfigure}
    \hfill
    \begin{subfigure}{0.4\textwidth}
        \centering
        	\tikzset{stroke/.style={black, line width=0.1mm}}

	\begin{tikzpicture}
\draw [stroke] (0.0,1.45) -- (0.02,1.44) -- (0.05,1.44) -- (0.24,1.44) -- (0.31,1.44) -- (0.36,1.44) -- (0.42,1.45) -- (0.49,1.45) -- (0.56,1.45) -- (0.62,1.45) -- (0.69,1.45) -- (0.76,1.45) -- (0.82,1.47) -- (0.87,1.47) -- (1.02,1.47) -- (1.07,1.49) -- (1.13,1.49) -- (1.15,1.49) ;
\draw [stroke] (0.05,1.36) -- (0.07,1.35) -- (0.16,1.29) -- (0.22,1.27) -- (0.25,1.24) -- (0.35,1.18) -- (0.4,1.16) -- (0.45,1.13) -- (0.51,1.09) -- (0.55,1.07) -- (0.6,1.04) -- (0.64,1.0) -- (0.69,0.98) -- (0.75,0.95) -- (0.8,0.91) -- (0.82,0.91) -- (0.84,0.91) -- (0.82,0.91) -- (0.82,0.89) -- (0.8,0.89) -- (0.71,0.85) -- (0.56,0.78) -- (0.51,0.76) -- (0.47,0.73) -- (0.42,0.71) -- (0.36,0.67) -- (0.22,0.6) -- (0.13,0.56) -- (0.11,0.55) -- (0.09,0.53) -- (0.09,0.55) -- (0.09,0.53) -- (0.07,0.53) -- (0.07,0.51) -- (0.05,0.51) -- (0.07,0.51) -- (0.11,0.49) -- (0.16,0.49) -- (0.15,0.49) -- (0.22,0.49) -- (0.27,0.49) -- (0.44,0.49) -- (0.49,0.49) -- (0.53,0.49) -- (0.64,0.49) -- (0.78,0.51) -- (0.91,0.51) -- (0.93,0.51) -- (0.96,0.51) -- (1.0,0.51) ;
\draw [stroke] (0.45,1.87) -- (0.47,1.87) -- (0.45,1.85) -- (0.45,1.8) -- (0.45,1.76) -- (0.47,1.78) -- (0.45,1.76) -- (0.45,1.73) -- (0.44,1.73) -- (0.45,1.73) -- (0.53,1.71) -- (0.56,1.71) -- (0.62,1.71) -- (0.73,1.73) -- (0.76,1.73) -- (0.8,1.75) -- (0.82,1.75) ;
\draw [stroke] (0.64,1.89) -- (0.65,1.87) -- (0.65,1.78) -- (0.65,1.75) -- (0.65,1.73) -- (0.65,1.69) -- (0.65,1.67) -- (0.65,1.62) -- (0.64,1.62) -- (0.65,1.56) -- (0.65,1.55) ;
\draw [stroke] (0.29,0.25) -- (0.31,0.25) -- (0.31,0.27) -- (0.33,0.31) -- (0.27,0.29) -- (0.25,0.31) -- (0.25,0.29) -- (0.18,0.24) -- (0.16,0.2) -- (0.13,0.09) -- (0.15,0.04) -- (0.18,0.05) -- (0.22,0.05) -- (0.25,0.15) -- (0.27,0.16) -- (0.27,0.2) -- (0.29,0.22) -- (0.31,0.22) -- (0.31,0.2) -- (0.33,0.09) -- (0.33,0.07) -- (0.36,0.02) -- (0.4,0.0) -- (0.42,0.0) ;
\draw [stroke] (0.55,0.27) -- (0.55,0.25) -- (0.56,0.25) -- (0.64,0.25) -- (0.65,0.25) -- (0.65,0.24) -- (0.71,0.25) -- (0.73,0.25) ;
\draw [stroke] (0.51,0.11) -- (0.53,0.11) -- (0.6,0.13) -- (0.67,0.15) -- (0.69,0.15) ;
\draw [stroke] (0.93,0.38) -- (0.93,0.36) -- (0.95,0.36) -- (0.95,0.35) -- (0.95,0.29) -- (0.96,0.25) -- (0.95,0.16) -- (0.95,0.13) -- (0.95,0.11) -- (0.95,0.07) -- (0.95,0.05) -- (0.95,0.04) -- (0.96,0.04) -- (0.95,0.02) -- (0.95,0.0) ;
\draw [stroke] (1.71,1.29) -- (1.69,1.29) -- (1.69,1.31) -- (1.64,1.29) -- (1.6,1.27) -- (1.56,1.25) -- (1.53,1.24) -- (1.49,1.2) -- (1.44,1.13) -- (1.4,1.07) -- (1.38,1.02) -- (1.36,0.98) -- (1.36,0.93) -- (1.36,0.87) -- (1.38,0.84) -- (1.4,0.8) -- (1.42,0.76) -- (1.44,0.75) -- (1.47,0.73) -- (1.51,0.73) -- (1.55,0.73) -- (1.6,0.75) -- (1.62,0.75) -- (1.64,0.76) -- (1.65,0.76) -- (1.65,0.78) -- (1.69,0.8) -- (1.71,0.8) ;
\draw [stroke] (2.11,0.82) -- (2.11,0.84) -- (2.11,0.89) -- (2.09,0.89) -- (2.09,0.91) -- (2.04,0.91) -- (2.02,0.91) -- (1.95,0.8) -- (1.91,0.76) -- (1.89,0.71) -- (1.89,0.67) -- (1.89,0.65) -- (1.89,0.62) -- (1.93,0.6) -- (1.96,0.6) -- (1.95,0.6) -- (1.96,0.6) -- (2.02,0.65) -- (2.04,0.69) -- (2.05,0.71) -- (2.07,0.75) -- (2.09,0.78) -- (2.11,0.78) -- (2.11,0.76) -- (2.11,0.69) -- (2.11,0.65) -- (2.13,0.6) -- (2.18,0.58) -- (2.2,0.58) -- (2.24,0.58) ;
\draw [stroke] (2.4,1.13) -- (2.42,1.13) -- (2.51,1.13) -- (2.53,1.13) -- (2.6,1.13) -- (2.65,1.13) -- (2.69,1.11) -- (2.73,1.11) -- (2.76,1.11) -- (2.82,1.11) -- (2.85,1.11) -- (2.91,1.11) -- (2.95,1.11) -- (2.98,1.11) -- (3.0,1.11) -- (3.04,1.11) -- (3.05,1.11) ;
\draw [stroke] (2.44,0.93) -- (2.44,0.91) -- (2.45,0.91) -- (2.53,0.91) -- (2.55,0.91) -- (2.65,0.91) -- (2.69,0.91) -- (2.78,0.91) -- (2.82,0.93) -- (2.87,0.93) -- (2.93,0.91) -- (2.96,0.91) ;
\draw [stroke] (3.36,1.42) -- (3.38,1.42) -- (3.45,1.42) -- (3.45,1.44) -- (3.55,1.42) -- (3.55,1.4) -- (3.58,1.35) -- (3.6,1.31) -- (3.6,1.27) -- (3.6,1.24) -- (3.6,1.13) -- (3.56,0.98) -- (3.55,0.93) -- (3.51,0.87) -- (3.47,0.82) -- (3.45,0.78) -- (3.42,0.73) -- (3.38,0.69) -- (3.35,0.67) -- (3.33,0.65) -- (3.31,0.64) -- (3.29,0.64) -- (3.31,0.64) -- (3.33,0.65) -- (3.4,0.67) -- (3.44,0.69) -- (3.51,0.71) -- (3.56,0.71) -- (3.58,0.71) -- (3.62,0.69) -- (3.64,0.69) ;
\draw [stroke] (3.96,1.31) -- (3.98,1.29) -- (3.98,1.27) -- (4.0,1.2) -- (4.0,1.16) -- (4.0,1.13) -- (3.98,1.07) -- (3.98,1.04) -- (3.96,1.0) -- (3.95,0.95) -- (3.95,0.91) -- (3.95,0.85) -- (3.95,0.82) -- (3.93,0.78) -- (3.93,0.76) -- (3.93,0.75) -- (3.93,0.71) -- (3.93,0.73) -- (3.95,0.8) -- (3.95,0.85) -- (3.95,0.93) -- (3.96,1.0) -- (3.96,1.04) -- (3.98,1.07) -- (4.0,1.11) -- (4.0,1.18) -- (4.02,1.25) -- (4.04,1.29) -- (4.05,1.33) -- (4.05,1.35) -- (4.07,1.36) -- (4.13,1.4) -- (4.13,1.42) -- (4.18,1.42) -- (4.18,1.44) -- (4.27,1.42) -- (4.29,1.4) -- (4.31,1.35) -- (4.31,1.31) -- (4.31,1.27) -- (4.29,1.24) -- (4.25,1.16) -- (4.2,1.11) -- (4.16,1.07) -- (4.13,1.05) -- (4.04,1.02) -- (4.02,1.02) -- (4.0,1.04) -- (3.98,1.02) -- (4.0,1.04) -- (4.07,1.07) -- (4.16,1.07) -- (4.16,1.09) -- (4.25,1.04) -- (4.27,1.04) -- (4.31,1.0) -- (4.35,0.93) -- (4.35,0.85) -- (4.35,0.8) -- (4.35,0.76) -- (4.35,0.78) -- (4.31,0.73) -- (4.31,0.71) -- (4.25,0.67) -- (4.24,0.67) -- (4.22,0.65) -- (4.2,0.65) -- (4.13,0.65) -- (3.98,0.71) -- (3.95,0.73) -- (3.87,0.75) ;
\draw [stroke] (4.56,1.09) -- (4.58,1.09) -- (4.58,1.11) -- (4.62,1.11) -- (4.75,1.11) -- (4.8,1.13) -- (4.84,1.13) -- (4.91,1.13) -- (4.91,1.15) -- (4.95,1.13) ;
\draw [stroke] (4.76,1.31) -- (4.76,1.29) -- (4.78,1.2) -- (4.8,1.15) -- (4.8,1.11) -- (4.78,1.04) -- (4.76,0.91) -- (4.76,0.89) -- (4.76,0.85) -- (4.76,0.84) -- (4.76,0.8) ;
\draw [stroke] (5.29,1.44) -- (5.29,1.42) -- (5.27,1.4) -- (5.25,1.31) -- (5.24,1.25) -- (5.22,1.22) -- (5.2,1.18) -- (5.18,1.15) -- (5.16,1.11) -- (5.15,1.09) -- (5.13,1.05) -- (5.11,1.04) -- (5.11,1.04) -- (5.13,1.0) -- (5.11,1.0) -- (5.13,1.0) -- (5.22,1.0) -- (5.27,1.02) -- (5.29,1.02) -- (5.33,1.04) -- (5.42,1.04) -- (5.44,1.05) -- (5.45,1.05) -- (5.51,1.05) -- (5.55,1.07) ;
\draw [stroke] (5.38,1.38) -- (5.36,1.36) -- (5.38,1.27) -- (5.38,1.24) -- (5.38,1.15) -- (5.36,1.05) -- (5.36,1.0) -- (5.36,0.96) -- (5.36,0.91) -- (5.36,0.87) -- (5.36,0.84) -- (5.35,0.8) -- (5.36,0.76) ;
\draw [stroke] (5.6,1.38) -- (5.62,1.38) -- (5.64,1.36) -- (5.71,1.36) -- (5.78,1.36) -- (5.82,1.36) -- (5.8,1.36) -- (5.82,1.36) -- (5.89,1.36) -- (5.96,1.36) -- (6.0,1.38) ;
\draw [stroke] (5.62,1.33) -- (5.62,1.31) -- (5.65,1.24) -- (5.65,1.18) -- (5.65,1.15) -- (5.64,1.07) -- (5.64,1.04) -- (5.64,1.0) -- (5.64,0.96) -- (5.64,0.93) -- (5.64,0.89) -- (5.62,0.84) -- (5.62,0.8) -- (5.62,0.75) -- (5.6,0.73) ;
\draw [stroke] (5.62,1.07) -- (5.64,1.07) -- (5.65,1.07) -- (5.75,1.07) -- (5.76,1.07) -- (5.8,1.09) -- (5.82,1.09) -- (5.84,1.09) -- (5.85,1.09) -- (5.85,1.07) ;

		   \end{tikzpicture}
        \caption{Ex.2 $ \sum_{a=1}^{4} C_a = 2B +4F$}
        \label{fig:attention:ex2}
    \end{subfigure}
    \begin{subfigure}{0.4\textwidth}
        \includegraphics[trim={60 60 60 60},clip,width=\linewidth]{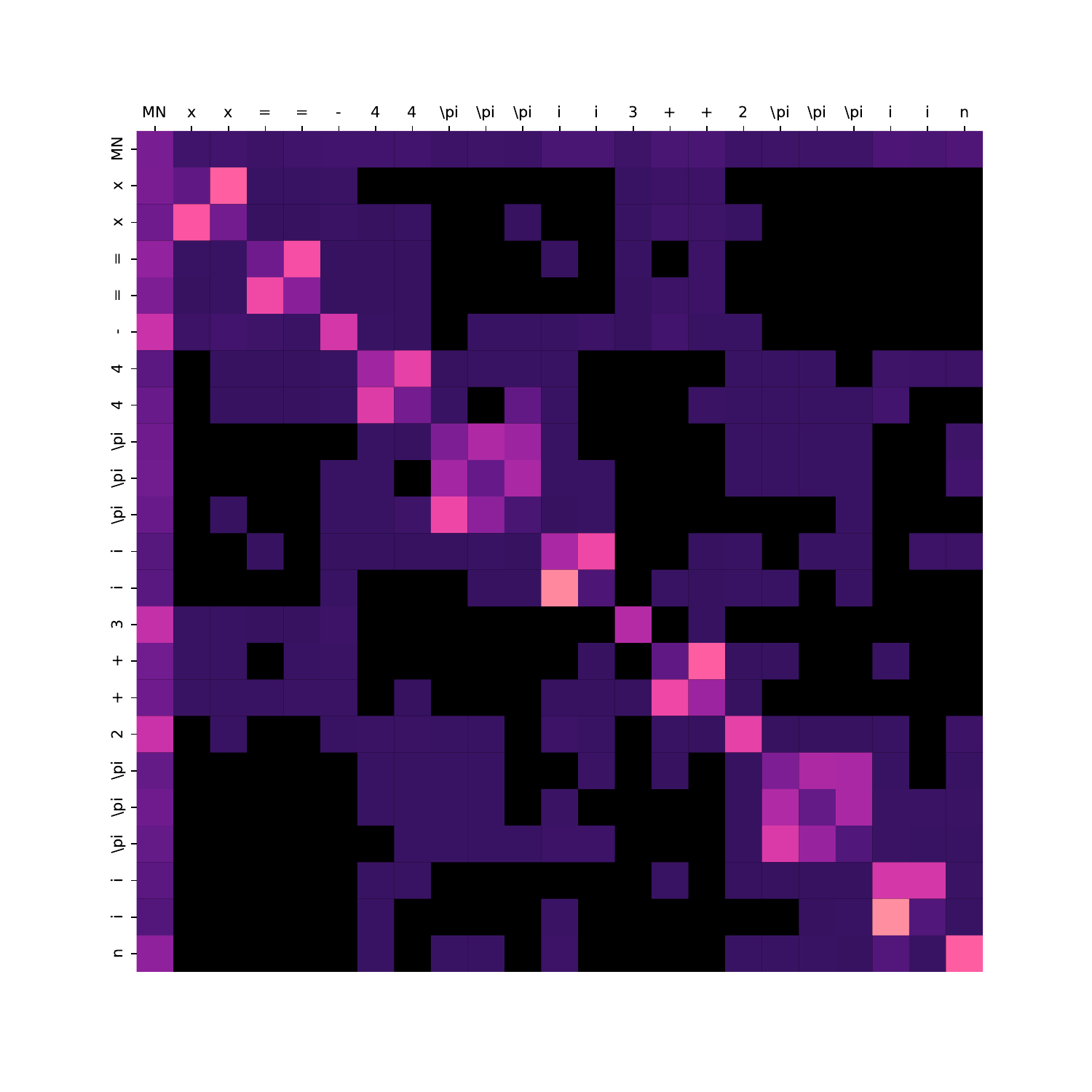}
        \caption{Attention visualization of Ex.1}
        \label{fig:attention_v:ex1}
    \end{subfigure}
    \hfill
    \begin{subfigure}{0.4\textwidth}
        \includegraphics[trim={60 60 60 60},clip,width=\linewidth]{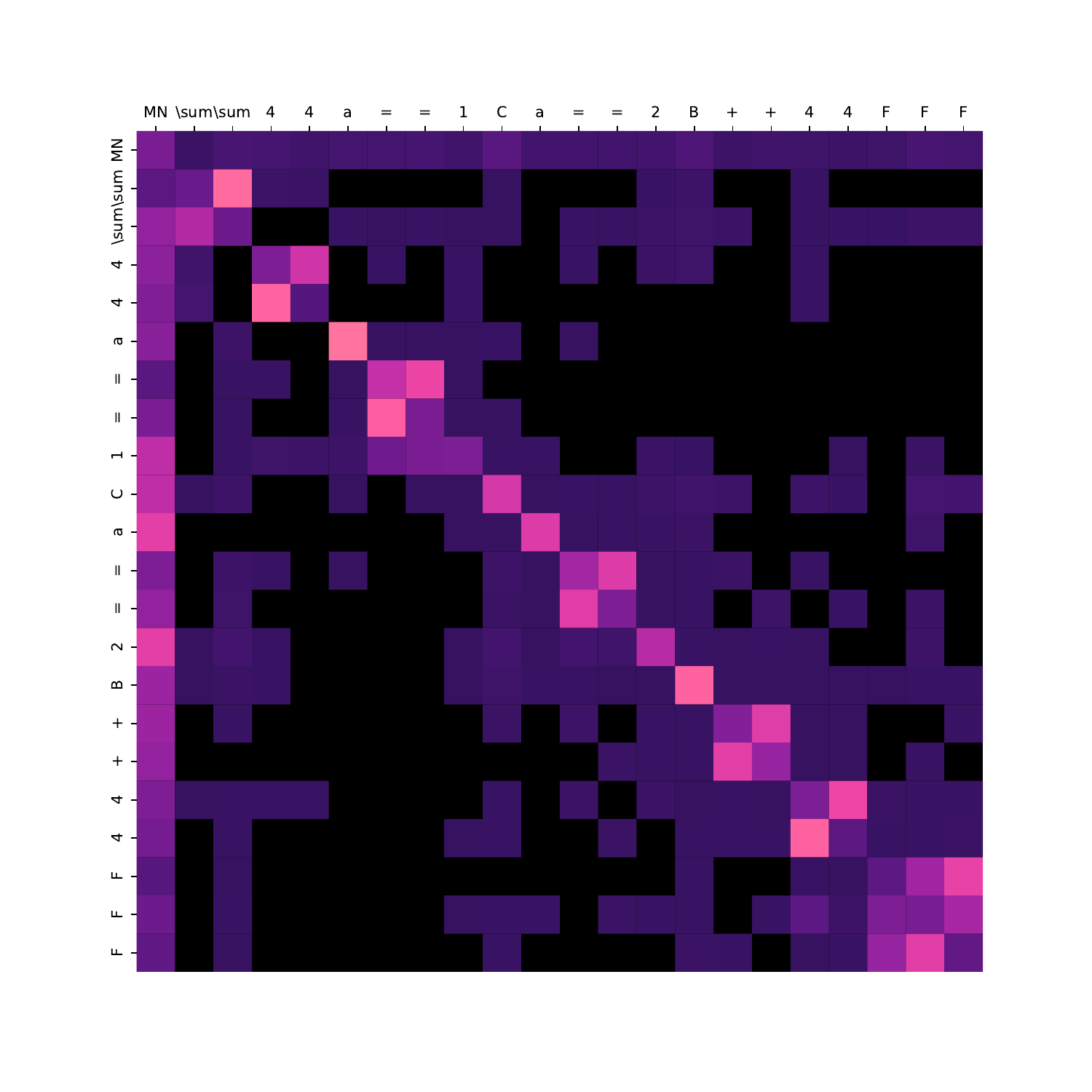}
        \caption{Attention visualization of Ex.2}
        \label{fig:attention_v:ex2}
    \end{subfigure}
    \caption{Attention Visualization Examples.
    The lighter color represents the higher attention score, which means the more important the neighbor node or the connected edge.
The darker color is the opposite, and the pure black means no connected edge between the nodes.}
    \label{fig:attention}
\end{figure}
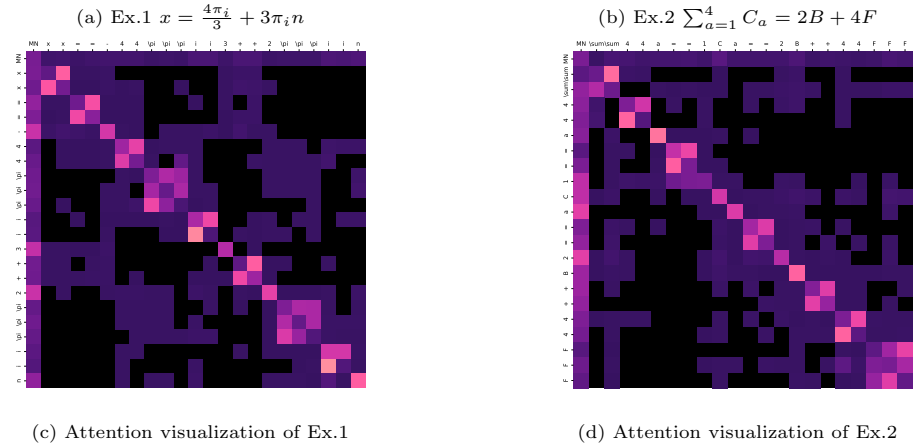

All the attention scores of one graph with $N$ nodes are saved in a $N\times N$ matrix, which is an adjacency matrix with attention scores as the edge weights.
We visualize the attention scores in heatmap, as shown in Fig. \ref{fig:attention_v:ex1} and \ref{fig:attention_v:ex2}.
The attention scores $\alpha_{ij}$ means the importance of the $i^{th}$ stroke to the $j^{th}$ stroke, which is the value of row $i$, column $j$ in heatmap.
The first node of each graph is the master node, the following nodes are the strokes by the order of writing.
We can obviously notice that the strokes belonging to the same symbol are more likely to have higher attention scores with each other, which means this stroke learn more information from the neighbors of the same symbol. 
Such as ``$\pi$'', ``$i$'', ``$x$ ''and ``$+$'' in Fig. \ref{fig:attention:ex1}, and also ``$\sum$'' and ``$F$'' in Fig. \ref{fig:attention:ex2}.
These performances demonstrate the model can effectively work on the segmentation of expression.

It is additionally interesting to observe the relationship between Master Node and other common nodes.
The first column of the heatmap is always full of light color, which means the master node obtains information from all the other nodes. 
While the first row is usually dark, only several strokes have a marginally lighter color, which means the global information of master node is more important to certain nodes, such as ``$C$'' and ``$B$'' in Fig. \ref{fig:attention:ex2}.
We conjecture that Mater Node contains global information that guides ``$C$'' and ``$B$'' to prioritize capital letters in their classification, which could have a significant positive impact on correct classification, since the capital letter and lowercase letter ``$C$'' are very similar in shape.

\subsubsection{Effect of Expression Length}
\label{analysis:complexity}
\begin{figure}[h]
    \centering
    \begin{subfigure}{0.48\textwidth}
        \includegraphics[width=\linewidth]{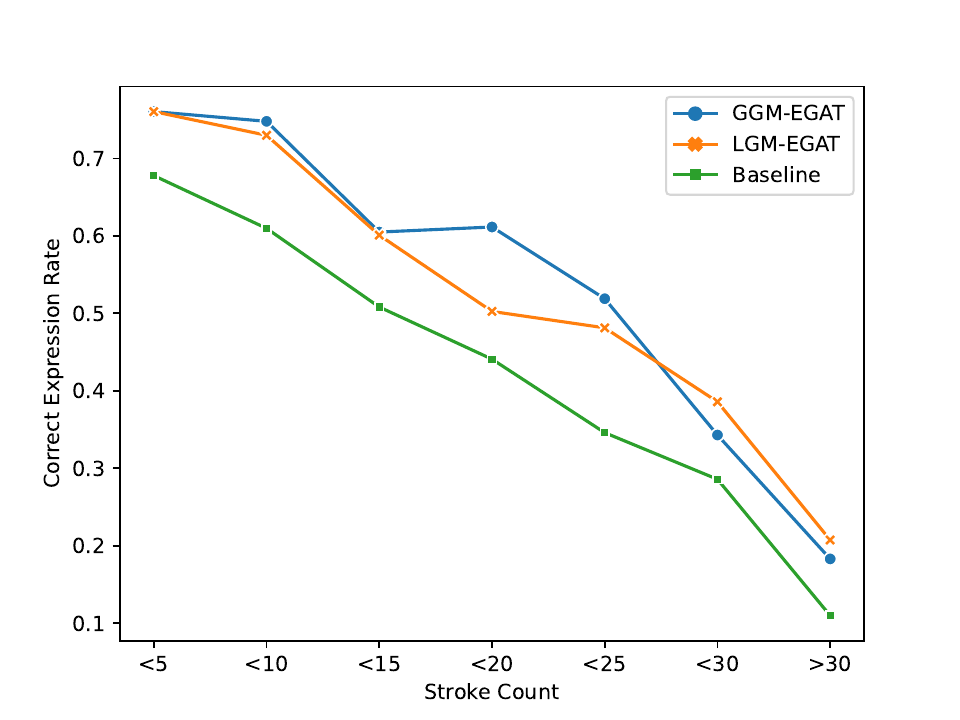}
        \caption{2019 Testset}
        \label{fig:complexity:2019_stroke}
    \end{subfigure}
    \hfill
    \begin{subfigure}{0.48\textwidth}
        \includegraphics[width=\linewidth]{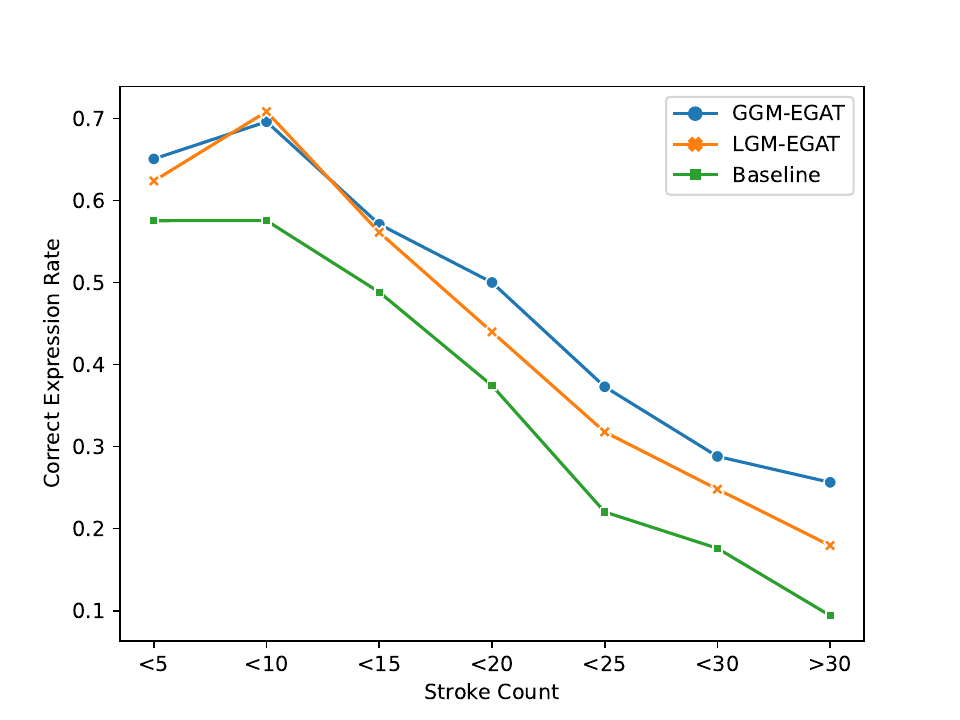}
        \caption{2023 Testset}
        \label{fig:complexity:2023_stroke}
    \end{subfigure}
    \begin{subfigure}{0.48\textwidth}
        \includegraphics[width=\linewidth]{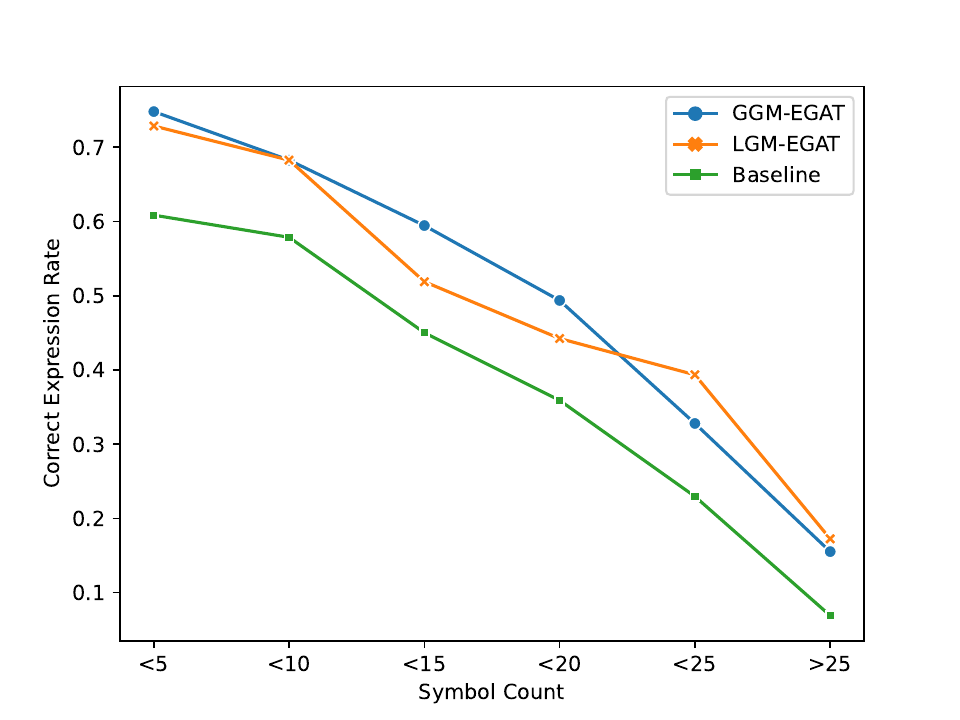}
        \caption{2019 Testset}
        \label{fig:complexity:2019_symbol}
    \end{subfigure}
    \hfill
    \begin{subfigure}{0.48\textwidth}
        \includegraphics[width=\linewidth]{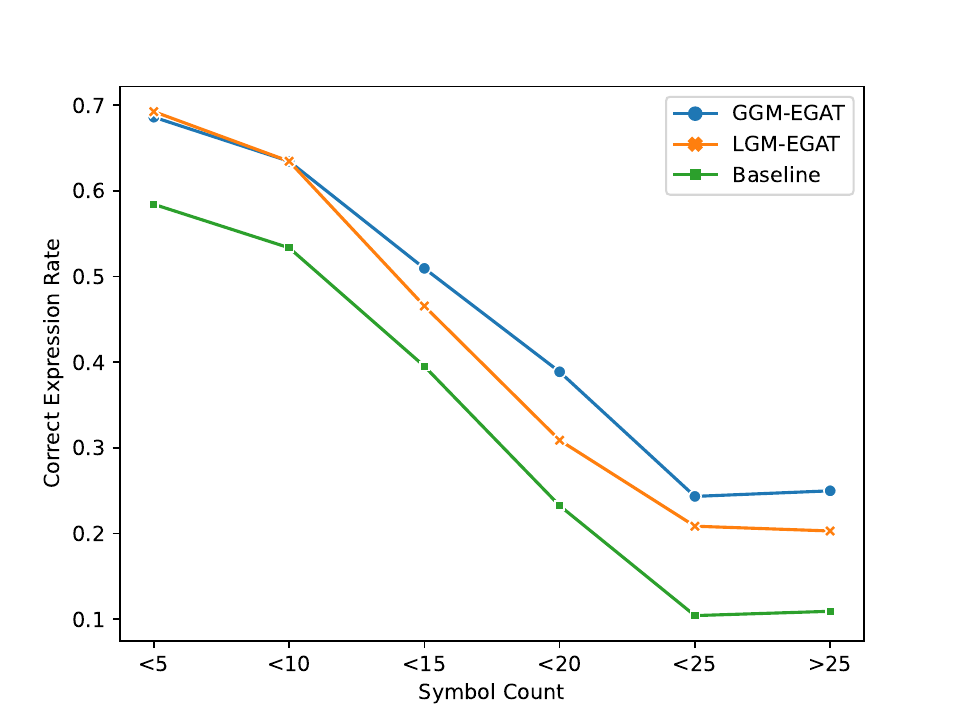}
        \caption{2023 Testset}
        \label{fig:complexity:2023_symbol}
    \end{subfigure}
    \caption{Effect of Expression Length.
    \ref{fig:complexity:2019_stroke} and \ref{fig:complexity:2023_stroke} are the correct expression rate of different expression lengths by strokes count.
    \ref{fig:complexity:2019_symbol} and \ref{fig:complexity:2023_symbol} are the file correct expression rate of different expression lengths by symbol count.}
    \label{fig:complexity}
\end{figure}

Expression Length can express the complexity of the math expression.
We plotted charts to analyze the file correct recognition rate of different expression lengths by strokes count and symbol count, as shown in Fig. \ref{fig:complexity}.
In general, the model performs better on shorter expressions, the recognition rate decreases as the expression length increases.
The proposed GGM-EGAT model and LGM-EGAT model have a significant improvement than baseline model, both in simple and complex expressions.
In addition, the GGM-EGAT model and LGM-EGAT model have a similar performance in simple expressions. 
The GGM-EGAT model with global information has better performance with the increase of expression complexity, especially on the 2023 dataset, which implies the significance of global information for the recognition of complex expressions.

\subsubsection{Error Analysis}
\label{analysis:error}
A summary of confusion histograms, detailing the most common errors in both individual symbols and symbol pairs, is conducted by the proposed GGM-EGAT model in the CROHME 2023 Test set.

\begin{table}[h]
    \centering
    \scalebox{0.8}{
    \begin{tabular}{c|c|cc|cc|cc|cc|cc}
    \hline
    & & \textbf{E} & \textbf{\#} & $\mathbf{err_1}$ & \textbf{\#}  & $\mathbf{err_2}$  &\textbf{\#}  & $\mathbf{err_3}$  &\textbf{\#} & $\mathbf{err_4}$  & \textbf{\#}  \\
    \hline
    \multirow{4}{*}{\textbf{Symbols}}&$1^{st} $ & $\mathit{1}$ & 133 & \textit{,} & 15 & \textit{(} & 13 & $\mathit{11}$ & 12 & $\mathit{\prime}$ & 10 \\
    &$2^{nd} $ & $\mathit{x}$ & 67 & $\mathit{X}$ & 24 & $\mathit{\times}$ & 8 & $\mathit{k}$ & 5 & $\mathit{u}$ & 5 \\
    &$3^{rd} $ & $\mathit{n}$ & 58 & $\mathit{m}$ & 20 & $\mathit{u}$ & 8 & $\mathit{4}$ & 4 & $\mathit{x}$ & 4 \\
    &$4^{th} $ & $\mathit{-}$ & 47 & $\mathit{1}$ & 4 & $\mathit{--}$ & 4 & $\mathit{\sqrt{\ }}$ & 3 & $\mathit{+}$ & 3 \\
    \hline
    &$1^{st} $ & $\textit{x})$ & 24 & \textit{X}) & 10 & $\mathit{x\parallel}$)& 3 & - & - & -& - \\
    \textbf{Symbol}&$2^{nd} $ & $\mathit{\frac{}{2}}$ & 22 & $\mathit{-2}$ & 3 & -&-&-&-&-&-\\
    \textbf{Pairs}&$3^{rd} $ & $(\textit{x}$& 18 & $(\mathit{X} $ & 9 & -&-&-&-&-&-\\
    &$4^{th} $ & $\mathit{\frac{1}{}}$ & 20 & $\mathit{1\parallel-}$ & 5 & $\mathit{\frac{2}{}}$ & 3 & -&-&-&- \\
    \hline
    \end{tabular}}
    \caption{Error Analysis of Symbols and Symbol Pairs with the proposed GGM-EGAT model in CROHME 2023 Testset.
    $\mathbf{E}$ is the symbol or symbol pair for recognition, $\mathbf{\#}$ is the number of occurrences in the 2023 test set. $\mathbf{err_1}$, $\mathbf{err_2}$, $\mathbf{err_3}$, $\mathbf{err_4}$ are the most confusing mistakes of this symbol.
    Only the occurrences greater than 3 are listed.
    ``$\parallel$'' in Symbol Pairs means missing the relation between 2 symbols.}
    \label{tab:error_hist}
\end{table}

The confusion histogram is presented in Table \ref{tab:error_hist}. This table highlights the symbols and symbol pairs that posed the top 4 greatest difficulty for recognition in the CROHME 2023 Test set.
According to errors of symbols, similar shape of different symbols most likely lead to misclassification, such as ``$x$'', ``$X$'' and ``$\times$''.
Contextual information from neighboring nodes and the aggregation of master nodes can potentially assist, but a slight number of mistakes are still occurring.
Due to the incorrect prediction of some edges in the model, even if all the symbols are all predicted correctly, the relations between the symbols may also cause the errors. Such as all ``$\parallel$'' in Symbol Pairs, basic ``Right'' relation was not successfully predicted, due to edges prediction without structural constraints.



\section{Conclusion}
\label{conclusion}
In this paper, we present a novel comprehensive approach to Online Handwritten Mathematical Expression Recognition (OHMER) based on Stroke-level Graph Modeling.

First, we proposed a stroke-level graph modeling method that captures both local and global graph features. This method converts the HMER problem into node and edge classification task, addressing the complexities of both node and edge classifications. 
Second, we introduced an end-to-end architecture designed to effectively fuse embedded node and edge features. This fusion allows for simultaneous prediction of both node attributes and edge attributes. This unified approach ensures a more cohesive understanding of the mathematical expression by combining the strengths of local and global graph information.
Finally, through the integration of stroke-level graph modeling and the end-to-end fusion architecture, our system achieves significant improvements in symbol detection (94.12\% in CROHME 2019, 93.21\% in CROHME 2023), respectively relation classification (94.49\% in CROHME 2019, 93.45\% in CROHME 2023), and respectively expression-level recognition(60.72\% in CR-OHME 2019, 55.30\% in CROHME 2023).

In the future, we will extend our work using the proposed baseline method by applying a ``Stroke to Sequence'' strategy to handle larger and more diverse datasets. We will also explore a ``Graph to Stroke'' approach for generating online handwritten mathematical expressions, aiming to further enhance performance. Beyond HMER, we plan to investigate new applications of our system in other domains with 2D graph-structured data, such as chemical notation recognition and musical score analysis. These efforts will demonstrate the versatility and adaptability of our method across various fields.

\bibliography{baseline}
\bibliographystyle{splncs04}



\end{document}